\documentclass[10pt,twocolumn,letterpaper]{article}

\usepackage[final]{cvpr}      % To produce the REVIEW version
\usepackage{url}
\usepackage{amsmath}
\usepackage{amsfonts} % <-- Add this
\usepackage{algorithm}
\usepackage{algpseudocode}
\usepackage{booktabs}
\usepackage{multirow}
\usepackage{graphicx}     % For scaling tables or including images (optional)
\usepackage{multirow}     
\usepackage{array}    % For the p{} column type
\usepackage{adjustbox}    % For resizing tables to fit width, if needed (optional)
\usepackage{subcaption}
\newcommand{\xmark}{\ding{55}}%
\usepackage{pifont}
\usepackage[labelsep=period]{caption}
\usepackage{tabularx} % add in preamble

\definecolor{cvprblue}{rgb}{0.21,0.49,0.74}
\usepackage[pagebackref,breaklinks,colorlinks,allcolors=cvprblue]{hyperref}

\def\paperID{21514} % *** Enter the Paper ID here
\def\confName{CVPR}
\def\confYear{2026}

\title{TestMate: Test-Time Domain Adaptation Aided by Lightweight Vision Foundation Model}

\author{Dimitrios Fotiou \and Vasileios Mygdalis \\
Aristotle University of Thessaloniki\\
{\tt\small pitas@csd.auth.gr}
\and Ioannis Pitas
}
\begin{document}
\maketitle
\begin{abstract}
Test-Time Domain Adaptation (TTDA) aims to adapt Deep Neural Networks to distribution shifts using only streaming, unlabeled test data in real time. Current methods for semantic segmentation tasks suffer from critical limitations. Entropy minimization techniques require costly backpropagation, risking catastrophic forgetting and producing noisy segmentation boundaries. Memory-bank methods, while backpropagation-free, exhibit slow adaptation, requiring numerous samples to converge and struggle to handle continuous domain shifts. We introduce TestMate, a novel, real-time, and backpropagation-free TTDA framework that overcomes these issues. TestMate leverages generalization capability of a lightweight Visual Foundation Model to guide the adaptation. We use a zero-shot instance segmentation YOLOv8-seg based model to generate unlabeled mask proposals for objects and their parts at multiple scales in real time. These proposals are fused with the primary model via a heuristic, size-ordered competitive scheme, where small, high-confidence regions dominate and refine predictions in surrounding larger, less certain areas. This paremeter-free mechanism enables immediate adaptation from the first frame, inherently avoids catastrophic forgetting and effectively preserves fine object details and boundaries, even for small objects. TestMate can be used as a standalone, efficient refinement module or seamlessly integrated into existing TTDA methods to significantly boost their performance. We demonstrate state-of-the-art results across two benchmark datasets, proving TestMate's effectiveness in three distinct adaptation tasks: TTDA, Source-Free Domain Adaptation (SFDA), and online-TTDA. Code is available. 
\end{abstract}

\section{Introduction}
A key challenge in computer vision is ensuring the performance of Deep Neural Network (DNN) models under data probability distribution shifts, which often arise 
\begin{figure}[ht!]
    \centering
    \includegraphics[width=0.85\linewidth]{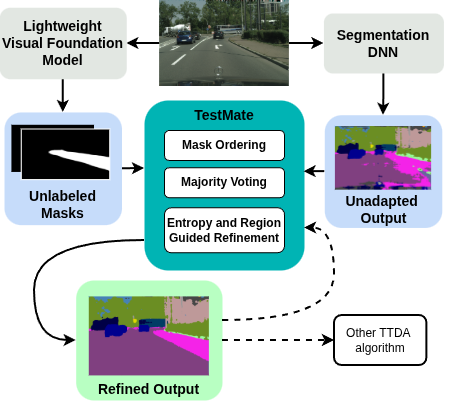}
    \caption{\fontsize{9}{11}\selectfont
        TestMate is a lightweight, backpropagation-free adaptation module. For a test image, a Segmentation DNN produces an initial output, while a Visual Foundation Model generates high-quality unlabeled masks. TestMate processes masks heuristically, from smallest to largest, labels them via majority voting, and fuses them with the DNN output, considering their entropy and region dominance. This parameter-free refinement produces a high-fidelity output with sharper boundaries even in small objects. TestMate can function standalone, be applied recursively, or enhance existing TTDA algorithms (dashed arrows).
    }
    \label{fig:model_overview_small}
\end{figure}
due to their limited generalization capabilities when trained on relatively small datasets. For example, autonomous vehicles need to operate robustly in adverse environmental conditions \cite{sakaridis2021acdc}, where illumination and visibility can be affected by heavy rain, snow, or fog. One strategy to improve DNN robustness is Domain Generalization, which involves training DNNs on large, diverse datasets \cite{rohlfs2025generalization}, often supplemented with synthetic data \cite{jaipuria2020deflating} that mimics the target data's probability distribution. However, because it is impractical to anticipate or simulate all real-world scenarios, there is growing research on Domain Adaptation (DA) methods that adapt pretrained models to new data distributions.

Domain adaptation aims to improve DNN performance when transitioning from a source data domain, used for training, to a target data domain. A widely studied case is Unsupervised Domain Adaptation (UDA) \cite{kouw2019review}, where target labels are unavailable during adaptation, but source data information is. However, in many real-world applications, access to source data may be restricted due to privacy, ownership, or storage constraints, leaving only the pretrained DNN model weights available. This more challenging setting is known as Source-Free Domain Adaptation (SFDA), which requires a DNN model to adapt without source data. In SFDA settings where target data also arrive sequentially during deployment, Test-Time Domain Adaptation (TTDA) must be employed. Furthermore, if the target data probability distribution changes continuously, the setting is referred to as online-TTDA. Since the boundaries between these settings are often inconsistently defined, we provide a summary in \cref{tab:domain_settings} for clarity. Our focus is on TTDA, but we also evaluate our approach in the SFDA and online-TTDA settings.

\begin{table}[h]
    \centering
    \fontsize{8}{11}\selectfont
    
    % Define a new column type 'C' that is centered and wraps (it's an 'X' column)
    \newcolumntype{C}{>{\centering\arraybackslash}X}
    
    % We use tabularx to make the table fit the \columnwidth
    % The 'X' columns will expand/wrap to fill the space
    % The column spec is: one 'l' column, and four 'C' (centered X) columns
    \begin{tabularx}{\columnwidth}{lCCCC}
        \toprule
        \textbf{Setting} & \textbf{Source Data} & \textbf{Target Labels} & \textbf{Online Streaming} & \textbf{Contin. Prob. Shift} \\
        \midrule
        Simple DA & Yes & Yes & No & No \\
        Unsupervised DA (UDA) & Yes & No & No & No \\
        Source-Free DA (SFDA) & No & No & No & No \\
        Test-Time DA (TTDA) & No & No & Yes & No \\
        Online Test-Time DA & No & No & Yes & Yes \\
        \bottomrule
    \end{tabularx}
    
    \caption{\fontsize{9}{11}\selectfont A Taxonomy of Domain Adaptation Settings: From Data Availability to Online Streaming and Continuous Distribution Shift}
    \label{tab:domain_settings}
\end{table}

Although TTDA methods have recently demonstrated strong performance on classification benchmarks like MNIST \cite{deng2012mnist} and SVHN \cite{netzer2011reading}, their effectiveness often diminishes on semantic segmentation tasks, such as the GTA-V \cite{richter2016playing}-to-Cityscapes \cite{cordts2016cityscapes} setting. Common TTDA methods, such as entropy minimization \cite{wang2020tent} and pseudo-labeling \cite{song2023ecotta}, are prone to noise and catastrophic forgetting. They also tend to focus disproportionately on dominant or low-entropy regions, often overlooking smaller or less frequent ones. Additionally, while DNN models may roughly localize objects, the resulting segmentation masks are frequently distorted, especially around object boundaries. Moreover, even gradient-free memory-bank methods that achieve more stable performance require substantial target data and multiple iterations to construct a reliable memory-bank. The number of iterations needed often scales with the source-target domain gap or the noise level of the model's output, delaying performance improvements. Consequently, their segmentation performance on initial samples is often low, limiting their suitability for critical TTDA applications that demand optimal performance from the very first sample.

Building on these insights, we explore Vision Foundation Models (VFM) as a potential solution to address the limitations in image segmentation inherent in standard TTDA methods. Pretrained on vast and varied image datasets with extensive augmentations, VFM excels at generalizing to unseen domains and capturing fine-grained object details. In our TTDA segmentation approach, namely TestMate, we first apply zero-shot model FastSAM \cite{zhao2023fast}, a lightweight VFM based on YOLOv8-seg \cite{jocher2023yolov8} model, to generate multiple coarse image segmentation proposals on each target-domain image in real time, even with limited GPU resources. We then fuse these proposals with the original DNN model outputs using a straightforward refinement algorithm that amplifies semantic segmentation on underrepresented classes, enhances their boundary precision, and mitigates noisy high-entropy regions. Our method is model-agnostic, can serve as an instance-based TTDA approach, or seamlessly integrate with any existing TTDA pipelines, providing less noisy input and boosting their accuracy and convergence.

The key contributions are as follows:
\begin{itemize}
    \item We introduce the use of a lightweight VFM to guide and refine TTDA for semantic segmentation.
    \item We propose TestMate, a parameter-free algorithm that enhances boundaries, improves small and rare object segmentation, and suppresses noise, effectively from the very first frame.
    \item TestMate is a model-agnostic, plug-and-play module that can serve as a standalone TTDA method or seamlessly integrate with existing TTDA pipelines to boost their performance.
    \item We demonstrate new state-of-the-art (SOTA) results across SFDA, TTDA, and online-TTDA segmentation benchmarks.
\end{itemize}

\section{Related Work}
\subsection{Test-Time Domain Adaptation}
Test-time domain adaptation methods can be categorized into data-based and model-based approaches \cite{li2024comprehensive}. Data-based TTDA methods aim to align the target domain data or its feature representations more closely with the source domain, without updating model parameters. For example, diffusion models trained on source data \cite{gao2023back, tsai2024gda}, inject noise into target data samples. During the denoising process, the reconstruction loss encourages a source-like performance style in the DNN inference. VDM-DA \cite{tian2021vdm} exploits a pre-trained model to create a virtual domain as a bridge between the source and target ones using Gaussian Mixture Modeling.

Another data-based strategy, class prototypes, defined as the mean feature vector of each class, are also utilized to align the source and target data probability distributions. For example, contrastive learning is employed to bring class prototypes, estimated during DNN training on source data,  closer to corresponding target features \cite{dobler2023robust}. A memory-bank, a collection of labels and features, is constructed during DNN model testing using low-entropy targets, allowing the estimation of class prototype features and labels to guide network adaptation \cite{wang2023feature}. Similarly, a semantic adaptation module is used in \cite{wang2023dynamically} that generates two types of class prototypes: one from the latest target data and another by averaging historical prototypes.

Model-based TTDA methods focus on fine-tuning the DNN model using target domain data. Since target labels are unavailable in TTDA scenarios, self-training is the most common approach for addressing this issue.  An entropy loss function is minimized to boost DNN model confidence \cite{wang2020tent, song2023ecotta}. Pseudo-labeling, which assigns DNN model class predictions directly from the model's output on target data \cite{dobler2023robust, ma2024improved, wang2022continual}, is another popular TTDA method. The model-based approaches may train the whole network, a slow process and prone to overfitting, only some layers, most commonly the batch normalization ones \cite{wang2020tent, kim2022ev}, or even domain-specific submodules \cite{song2023ecotta}.

Test-time domain adaptation methods typically update DNNs on-the-fly during deployment, making gradient-based approaches prone to overfitting and unstable training. A common remedy is the teacher–student framework \cite{wang2022continual}, where the student is trained on strongly augmented target images using pseudo-labels from the teacher. The student is updated frequently, while the teacher is updated more conservatively and robustly via an exponential moving average of the student weights. However, such methods often require multiple training rounds to converge. Similarly, prototype-based approaches rely on numerous target samples to construct reliable memory-banks. To address these limitations, dynamic update frequency strategies have been proposed for the teacher–student framework \cite{zhao2023towards} that reduce the time needed for convergence, but still is unable to provide a significant accuracy boost increase from the first frame.

% Recent TTDA methods have achieved strong results, but they are mainly tailored for classification tasks. Gradient-based approaches \cite{song2023ecotta, zhao2023towards} offer stable TTDA performance through teacher-student frameworks. Gradient-free methods \cite{wang2023dynamically} are efficient and perform closely to the gradient-based ones. However, current state-of-the-art TTDA methods often fail to accurately capture segmented object boundaries. In particular, small objects are frequently missed due to the lack of external information about the object region and content structure. Gradient-based TTDA methods require several adaptation steps, and gradient-free approaches need multiple target domain data samples to build accurate memory-banks. As a result, both are unreliable during the early adaptation stages and are unsuitable for image segmentation applications that require good-quality region masks from the first frame.

\subsection{Foundation Models in Domain Adaptation}
While Foundation Models (FM) have been extensively used in multiple areas of computer vision, their application in TTDA is limited. In \cite{tang2024source}, the authors proposed adapting both a vision-language foundation model and a source model through prompting and knowledge distillation but for classification purposes and without exploiting lightweight FMs that fulfill the real-time requirements of the TTDA task. Similarly, a powerful VFM, the Segment Anything Model (SAM) \cite{kirillov2023segment}, was leveraged in \cite{yan2023sam4udass} within a self-training framework for unsupervised domain adaptation in intelligent vehicle scenarios, but it incorporates source-domain information and thus cannot be applied in TTDA.

To the best of our knowledge, no prior work has investigated the use of Vision Foundation Models for TTDA in semantic segmentation. Overall, TTDA presents unique challenges, as adaptation must occur on the fly during deployment, demanding TTDA methods that are both accurate and computationally efficient. As a result, large foundation models such as CLIP \cite{radford2021learning} or SAM \cite{kirillov2023segment} may be impractical for real-world TTDA applications, due to their high resource requirements and computational latency.

\section{Method}
\subsection{Problem Setup}
Let \(f_{\boldsymbol{\theta}}:\mathcal{X}\to\mathcal{Y}\) be a semantic image segmentation DNN trained on a labeled dataset from a source domain \(\mathcal{D}_S\). At test time, it is deployed on an unseen target domain \(\mathcal{D}_T\), where samples arrive sequentially as \(\{\mathbf{X}_t\}_{t=1}^T\). Owing to the data probability distribution shift between \(\mathcal{D}_S\) and \(\mathcal{D}_T\), the DNN model's logits predictions \(\mathbf{\hat{Y}}_t = f_{\boldsymbol{\theta}}(\mathbf{X}_t) \in \mathbb{R}^{C \times H \times W}\) typically exhibit high entropy, causing the predicted segmentation mask \(\mathbf{\hat{Y}}_t' = \arg\max_{c} (\mathbf{\hat{Y}}_t) \in \mathbb{R}^{H \times W}\) to contain noisy regions and distorted object shapes.

In TTDA for semantic segmentation, each incoming image sample \(\mathbf{X}_t\) is used to dynamically adjust the behavior of \(f_{\boldsymbol{\theta}}\) on the fly, so as to reduce the expected image segmentation error on \(\mathcal{D}_T\). Crucially, this adaptation must operate online, processing each \(\mathbf{X}_t\) under realistic constraints on memory and computation.

\subsection{Overview}
Given a target input $\mathbf{X}_t$, the initial prediction $\mathbf{\hat{Y}}_t' = f_{\boldsymbol{\theta}}(\mathbf{X}_t)$ often exhibits high entropy and distorted boundaries due to the domain shift. To mitigate this, we introduce TestMate, a refinement module that leverages a lightweight VFM to produce cleaner, more accurate outputs. 

Specifically, we use a zero-shot instance segmenter, FastSAM \cite{zhao2023fast} (based on YOLOv8-Seg \cite{jocher2023yolov8}), as a small VFM. Although lightweight, FastSAM \cite{zhao2023fast} is pretrained on large, diverse datasets, enabling it to generalize across domains and produce high-quality, unlabeled masks for objects and their parts at multiple scales in real time.

TestMate exploits this robust generalization robustness. It utilizes the VFM's multi-scale masks as unlabeled spatial priors. It then intelligently fuses these spatial priors with the unadapted model's semantic logits through a parameter-free, heuristic-guided process. This refines the predictions, correcting noisy regions and sharpening object boundaries, without any gradient updates.

\subsection{TestMate}

At test time, given a target image \(\mathbf{X}_t\), in addition to obtaining the raw prediction logits \(\mathbf{\hat{Y}}_t = f_{\boldsymbol{\theta}}(\mathbf{X}_t) \in \mathbb{R}^{C \times H \times W}\) from the segmentation model, we extract a set of binary, unlabeled masks \(\{\mathbf{M}_i\}_{i=1}^N\) using a VFM \(g_{\boldsymbol{\phi}}\). Formally,  
\[
\{\mathbf{M}_i\}_{i=1}^N = g_{\boldsymbol{\phi}}(\mathbf{X}_t), \quad \mathbf{M}_i \in \{0,1\}^{H \times W},
\]  
where \(\boldsymbol{\phi}\) denotes the parameters of the foundation model and \(N\) is the number of object masks generated.

\begin{figure}[t]
  \centering
  \begin{subfigure}[t]{0.28\columnwidth}
    \includegraphics[width=\linewidth]{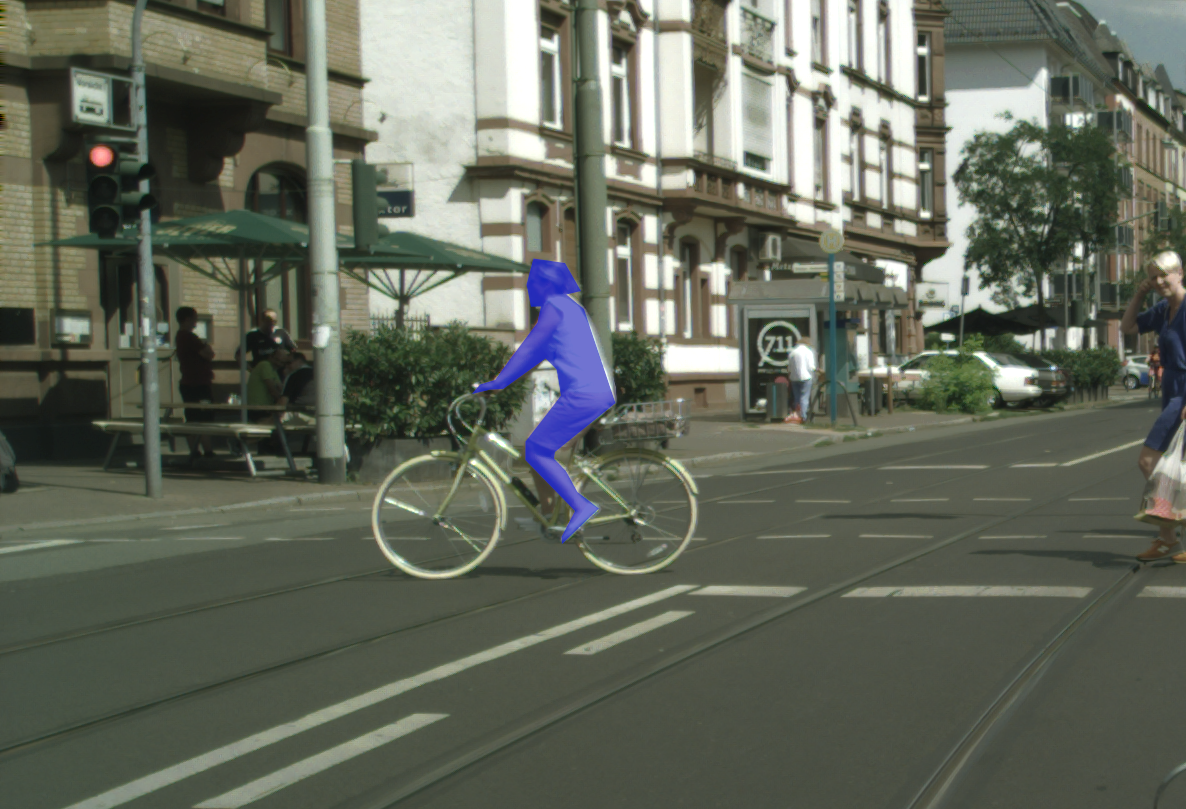}
    \caption{}
  \end{subfigure}
  \begin{subfigure}[t]{0.28\columnwidth}
    \includegraphics[width=\linewidth]{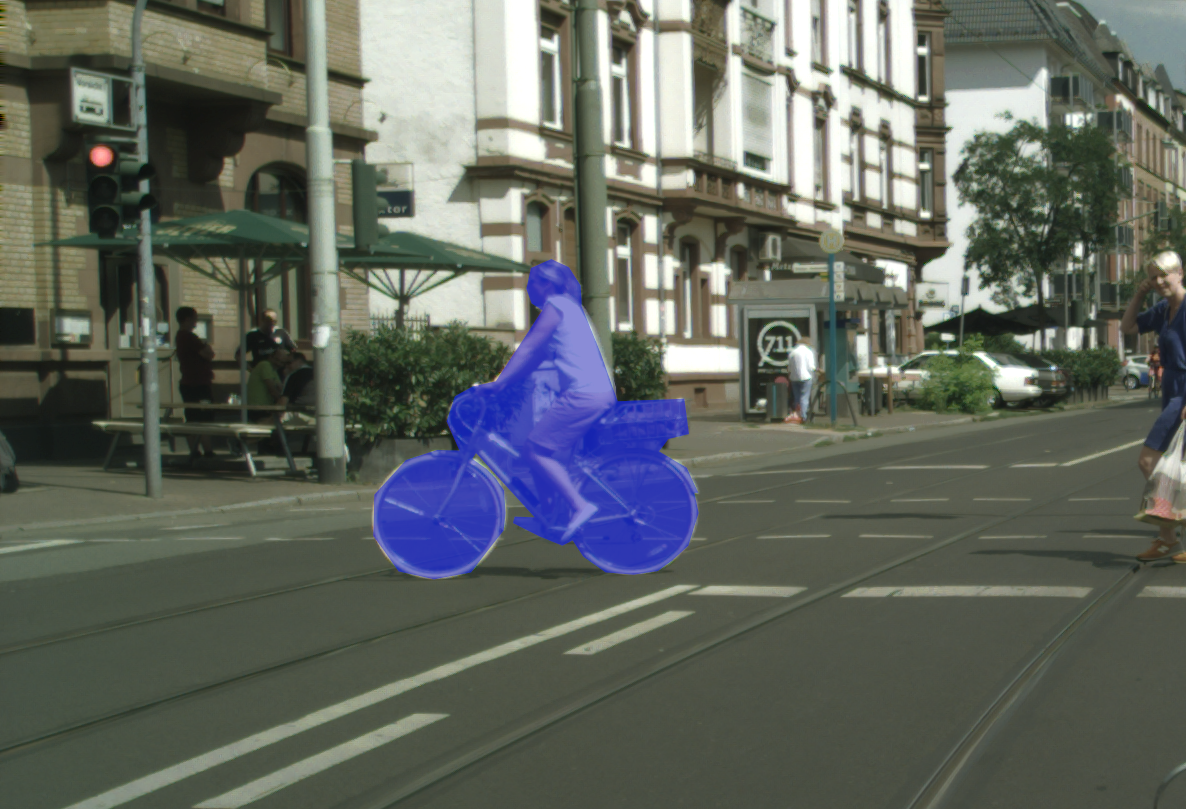}
    \caption{}
  \end{subfigure}
  \caption{\fontsize{9}{11}\selectfont Example cases of VFM Masks: (a) Single-object mask capturing a person; (b) Multi-object mask capturing both person and bike.}
  \label{fig:vfm_img_cat}
\end{figure}

We then identify the dominant class within a mask \(\mathcal{M}_i \subseteq \mathbf{\hat{Y}}_t\), where  
\[
\mathcal{M}_i = \{\, p \mid \mathbf{M}_i(p) = 1 \,\}
\]  
denotes the spatial region in the model’s output defined by the binary mask \(\mathcal{M}_i\) produced by the foundation model. The dominant class is determined by selecting the class that attains the highest total confidence over all pixels in the mask.
\begin{equation}
c_{\text{dom}} = \arg\max_{c} \sum_{p \in \mathcal{M}_i} 
\mathbb{I}\!\left(\arg\max_{c'} \mathbf{\hat{Y}}_t(p,c') = c\right),
\end{equation}
where \(\mathbf{\hat{Y}}_t(p,c)\) denotes the logit of class \(c\) at pixel \(p\), and \(\mathbb{I}(\cdot)\) is the indicator function.  
Equivalently, \(c_{\text{dom}}\) is the class that appears most frequently in the segmentation mask \(\mathbf{\hat{Y}}'_t = \arg\max_{c} \mathbf{\hat{Y}}_t\) restricted to \(\mathcal{M}_i\).  

We then partition \(\mathcal{M}_i\) into the dominant sub-region \(\mathcal{R}_i\) and the non-dominant sub-region \(\mathcal{N}_i = \mathcal{M}_i \setminus \mathcal{R}_i\).

The extracted unlabeled masks typically fall into two primary categories, illustrated in \cref{fig:vfm_img_cat}: (a) single-object masks, which accurately capture a single valid object or part of it, and (b) composite masks, which contain multiple valid objects or object parts within a single binary mask.

The simplest refinement strategy assigns class \(c_{\text{dom}}\) uniformly across \(\mathcal{M}_i\).  
For masks of category (a), typically corresponding to common object classes (e.g., humans, cars, trees, traffic signs, roads), this suppresses the noisy sub-region \(\mathcal{N}_i\). However, for composite masks (b), such naive reassignment suppresses smaller classes and distorts the shapes of objects. To address this limitation, we introduce three complementary refinement strategies.

\subsubsection{Ordered Refinement} The masks ${\mathcal{M}_i}$ are sorted by size in ascending order, with smaller masks processed first. Each image region is refined at most once. Smaller masks, which typically correspond to individual objects, are refined first. Larger masks, which may include multiple objects, are refined later, but previously updated smaller regions are excluded from further refinement.

\begin{table*}[ht!]
\centering
\footnotesize
\setlength{\tabcolsep}{1.5pt} % tighten horizontal spacing
\renewcommand{\arraystretch}{0.9} % tighten vertical spacing
\fontsize{9}{11}\selectfont
\begin{adjustbox}{max width=\textwidth}
\begin{tabular}{l|c|cccccccccccccccccccc}
\toprule
Method & BP & road & sidewalk & bilding & wall & fence & pole & light & sign & vege. & terrain & sky & person & rider & car & truck & bus & train & mbike & bike & mIoU \\
\midrule
Source model & - & 63.3 & 23.6 & 69.3 & 14.8 & 14.0 & 25.1 & 27.1 & 14.1 & 80.6 & 39.7 & 69.7 & 54.1 & 30.5 & 73.7 & 18.3 & 22.4 & 6.4 & 25.1 & 26.6 & 36.7 \\
\midrule
\multicolumn{22}{c}{SFDA} \\
\midrule
URMDA (CVPR’21) \cite{fleuret2021uncertainty} & \checkmark & 92.3 & 55.2 & 81.6 & 30.8 & 18.8 & 37.1 & 17.7 & 12.1 & 84.2 & 35.9 & 83.8 & 57.7 & 24.1 & 81.7 & 27.5 & 44.3 & 6.9 & 24.1 & 40.4 & 45.1\\
SFDA (CVPR’21) \cite{liu2021source} & \checkmark & 91.7 & 52.7 & 82.2 & 28.7 & 20.3 & 36.5 & 30.6 & 23.6 & 81.7 & 35.6 & 84.8 & 59.5 & 22.6 & 83.4 & 29.6 & 32.4 & 11.8 & 23.8 & 39.6 & 45.8 \\
SDF (MM’21) \cite{ye2021source} & \checkmark & \textbf{95.2} & 40.6 & \textbf{85.2} & 30.6 & 26.1 & 35.8 & 34.7 & 32.8 & 85.3 & 41.7 & 79.5 & 61.0 & 28.2 & 86.5 & 41.2 & \textbf{45.3} & 15.6 & 33.1 & 40.0 & 49.4 \\
HCL (NIPS’21) \cite{huang2021model} & \checkmark & 92.0 & 55.0 & 80.4 & 33.5 & 24.6 & 37.1 & 35.1 & 28.8 & 83.0 & 37.6 & 82.3 & 59.4 & 27.6 & 83.6 & 32.3 & 36.6 & 14.1 & 28.7 & 43.0 & 48.1 \\
DT-ST (CVPR’23) \cite{zhao2023towards} & \checkmark & 90.3 & 47.8 & 84.3 & \textbf{38.8} & 22.7 & 32.4 & 41.8 & 41.2 & 85.8 & 42.5 & 87.8 & 62.6 & 37.0 & 82.5 & 25.8 & 32.0 & 29.8 & \textbf{48.0} & 56.9 & 52.1 \\
TestMate(Ours) + DT-ST & \checkmark & 87.8 & 46.0 & 83.7 & 26.5 & \textbf{24.9} & \textbf{39.4} & \textbf{47.0} & \textbf{53.7} & \textbf{87.2} & \textbf{43.3} & \textbf{89.6} & \textbf{66.9} & \textbf{39.2} & \textbf{88.2} & \textbf{43.4} & 34.97 & \textbf{31.7} & 47.71 & \textbf{61.6} & \textbf{54.9} \\
\midrule
\multicolumn{22}{c}{TTDA} \\
\midrule
BSAdapt  & \xmark & 63.3 & 23.6 & 69.3 & 14.7 & 14.0 & 25.1 & 27.1 & 14.1 & 80.57 & 39.6 & 69.7 & 54.1 & 30.5 & 73.7 & 18.3 & 22.3 & 6.4 & 25.0 & 26.6 & 36.3 \\
TENT (ICLR'21) \cite{wang2020tent}  & \checkmark & 68.5 & 24.5 & 71.1 & 18.8 & 12.8 & 24.1 & 27.5 & 12.2 & 81.0 & 41.4 & 70.9 & 53.7 & 29.7 & 76.4 & 21.0 & 21.9 & 2.5 & 25.8 & 23.0 & 37.2 \\
EATA (ICML'22) \cite{niu2022efficient}  & \checkmark & 68.0 & 24.5 & 70.9 & 18.3 & 12.9 & 24.3 & 27.5 & 12.5 & 81.0 & 41.2 & 70.77 & 53.8 & 29.9 & 76.1 & 20.7 & 22.0 & 3.0 & 25.7 & 23.5 & 37.3 \\
CoTTa (CVPR'22) \cite{wang2022continual} &  \checkmark & 64.7 & 24.5 & 68.9 & 14.8 & 13.5 & 23.8 & 27.3 & 13.7 & 80.7 & 39.8 & 70.5 & 53.9 & 30.4 & 73.8 & 18.2 & 22.3 & 6.6 & 25.1 & 26.4 & 36.8 \\ 
%IN & \xmark  & &  &  &  &  &  &  &  &  &  &  & &  &  &  &  & &  &  & \\
%Momentum & \xmark &  &  &  &  &  &  &  &  &  &  &  & &  &  &  &  & &  &  & \\
DIGA (CVPR'23) \cite{wang2023dynamically} & \xmark & 83.2 & \textbf{28.9} & \textbf{79.7} & 16.7 & 21.2 & 27.4 & 33.9 & 30.2 & 82.3 & 36.4 & 71.6 & 55.1 & 17.2 & 80.0 & 26.0 & 32.0 & \textbf{4.7} & 22.1 & 36.6 & 41.3 \\
TestMate (Ours) & \xmark & 72.6 & 24.7 & 72.8 & \textbf{23.3} & 13.1 & 25.6 & 27.6 & 14.9 & 82.3 & \textbf{45.1} & \textbf{74.6} & \textbf{58.0} & \textbf{33.4} & 80.5 & 23.7 & 23.6 & 1.9 & \textbf{28.1} & 27.1 & 39.6 \\
TestMate (Ours) + CoTTA& \checkmark & 77.8 & 27.2 & 71.3 & 23.2 & 10.3 & 25.9 & 28.5 & 14.7 & \textbf{82.4} & 43.8 & 75.1 & 56.8 & 32.6 & 79.9 & 19.9 & 23.0 & 3.6 & 25.3 & 28.0 & 39.8 \\
TestMate (Ours) + DIGA  & \xmark & \textbf{85.1} & 24.9 & 79.4 & 19.3 & \textbf{23.8} & \textbf{27.5} & \textbf{33.2} & \textbf{30.9} & 81.7 & 40.3 & 72.1 & 55.6 & 17.6 & \textbf{81.0} & \textbf{28.4} & \textbf{34.4} & 4.3 & 23.3 & \textbf{36.8} & \textbf{42.2} \\
\bottomrule
\end{tabular}
\end{adjustbox}
\captionof{table}{\fontsize{9}{11}\selectfont Experimental results for GTA-V \cite{richter2016playing} → Cityscapes \cite{cordts2016cityscapes} (val) in SFDA and TTDA settings with DeepLabV2 \cite{chen2017deeplab} (ResNet-101 \cite{ he2016deep} backbone). BP denotes if the method utilizes backpropagation to achieve adaptation.}
\label{tab:miou-results_gta5-cityscapes}
\end{table*}

\subsubsection{Soft Refinement} Instead of hard refinement, we implement a weighted blending strategy that interpolates original logits and dominant-class logits. Given $\lambda_i = \frac{|\mathcal{R}_i|}{|\mathcal{M}_i|}$, the ratio of pixels in dominant region $\mathcal{R}_i$ compared to total pixels of $\mathcal{M}_i$, refined logits for pixel $p\in \mathcal{M}_i$ are computed as:
\begin{equation}
\hat{\mathbf{z}}_p = \mathrm{softmax}\!\left(
\lambda_i \cdot \bar{\mathbf{z}}_{\text{dom}} + (1 - \lambda_i) \cdot \mathbf{z}_p
\right),
\end{equation}

where $\bar{\mathbf{z}}_{\text{dom}}$ denotes the average logits of $\mathcal{R}_i$. This approach preserves uncertainty while reinforcing confident predictions. The impact of refinement is stronger when $\mathcal{R}_i$ represents a larger portion of the total pixels in $\mathcal{M}_i$, and when the entropy of the mean logit is low. In contrast, more ambiguous regions, those with a small dominant area or high entropy, contribute little to the refinement process.

\subsubsection{Entropy-based Filtering} We selectively refine uncertain pixels using an entropy criterion. Mean entropy within $\mathcal{R}_i$ is computed:

\begin{align}
\bar{H}_{\mathcal{R}_i} &= \frac{1}{|\mathcal{R}_i|} \sum_{p \in \mathcal{R}_i} H(p), \\
\text{where} \quad H(p) &= -\sum_c P(c \mid p) \log P(c \mid p),
\end{align}

Pixels in \(\mathcal{N}_i\) are refined only if their entropy exceeds this threshold:
\begin{equation}
H(p) > \bar{H}_{\mathcal{R}_i}, \quad \forall p \in \mathcal{N}_i.
\end{equation}

This targeted refinement approach preserves confident predictions  in non-dominant area of VFM's mask and exclude them from the refinement process.

\begin{algorithm}[ht!]
\caption{\fontsize{9}{11}\selectfont TestMate: VFM-Guided Test-Time Mask Refinement for TTDA Segmentation}
\label{alg:ttda_vfm}
\begin{algorithmic}[1]
\Require Target image $x_t$, segmentation model $f_\theta$, vision foundation model $g_\phi$
\State $\hat{y}_t, z_t \gets f_\theta(x_t)$ \Comment{Raw predictions and logits}
\State $\{\mathcal{M}_i\} \gets g_\phi(x_t)$ \Comment{Generate VFM masks}
\State Sort $\{\mathcal{M}_i\}$ by ascending size
\State $\mathcal{C} \gets \emptyset$ \Comment{Cache of already-refined pixels}
\For{each mask $\mathcal{M}_i$}
    \State $c_{\text{dom}} \gets \arg\max_c \sum_{p \in \mathcal{M}_i} \mathbb{I}(\hat{y}_p = c)$
    \State $\mathcal{R}_i \gets \{p \in \mathcal{M}_i \mid \hat{y}_p = c_{\text{dom}}\}$ \hspace{-2em} \Comment{Dominant region}
    \State $\mathcal{N}_i \gets \mathcal{M}_i \setminus \mathcal{R}_i$ \Comment{Non-dominant region}
    \State $\lambda \gets \frac{|\mathcal{R}_i|}{|\mathcal{M}_i|}$
    \State $\bar{z}_{\text{dom}} \gets \frac{1}{|\mathcal{R}_i|} \sum_{p \in \mathcal{R}_i} z_p$
    \State $\bar{H}_{\mathcal{R}_i} \gets \frac{1}{|\mathcal{R}_i|} \sum_{p \in \mathcal{R}_i} H(p)$
    \For{$p \in \mathcal{M}_i$}
        \If{$p \notin \mathcal{C}$ \textbf{and} ($p \in \mathcal{R}_i$ \textbf{or} $H(p) > \bar{H}_{\mathcal{R}_i}$)}
            \State $\hat{z}_p \gets \text{softmax}(\lambda \bar{z}_{\text{dom}} + (1{-}\lambda) z_p)$
            \State $\mathcal{C} \gets \mathcal{C} \cup \{p\}$ \Comment{Mark pixel as refined}
        \EndIf
    \EndFor
\EndFor
\end{algorithmic}
\end{algorithm}

Finally, the TestMate algorithm, illustrated in  \cref{alg:ttda_vfm}, can be applied optionally multiple times to each image. The VFM masks are generated once per image, after which TestMate can optionally refine the output masks recursively. By ascending ordering the VFM masks, applying soft refinements, and using entropy-based filtering, the algorithm effectively preserves small semantic classes, even when they are adjacent to larger semantic objects.

\begin{figure*}[ht!]
    \centering
    \setlength{\tabcolsep}{2pt}
    \renewcommand{\arraystretch}{0.7}
    \begin{tabular}{c c c c c c c}
        & \textbf{RGB} & \textbf{GT} & \textbf{Source} & \textbf{TENT} & \textbf{DIGA} & \textbf{TestMate} \\
        \raisebox{1\height}{\rotatebox{90}{\textbf{CS}}} &
        \includegraphics[width=0.12\textwidth]{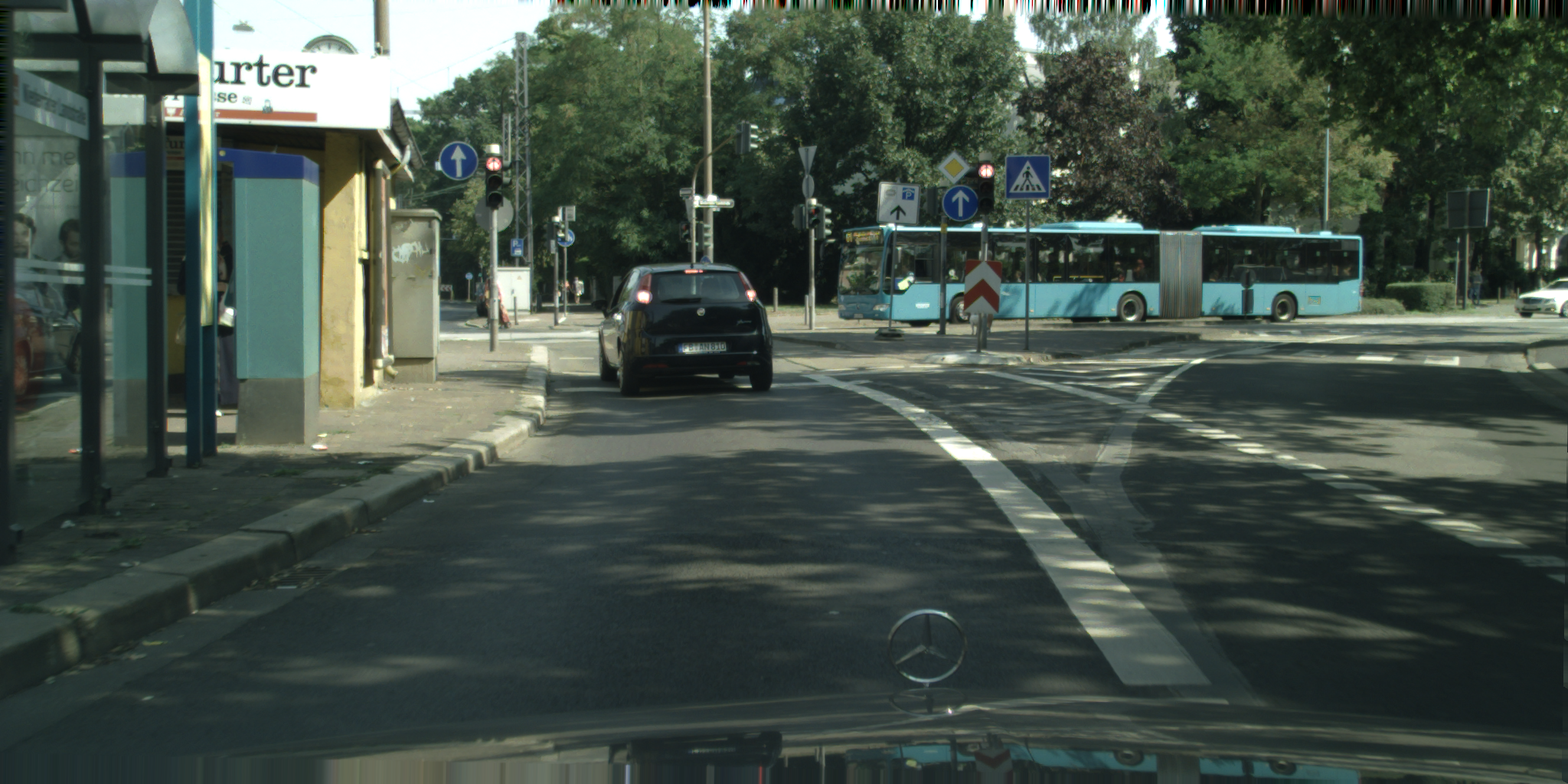} &
        \includegraphics[width=0.12\textwidth]{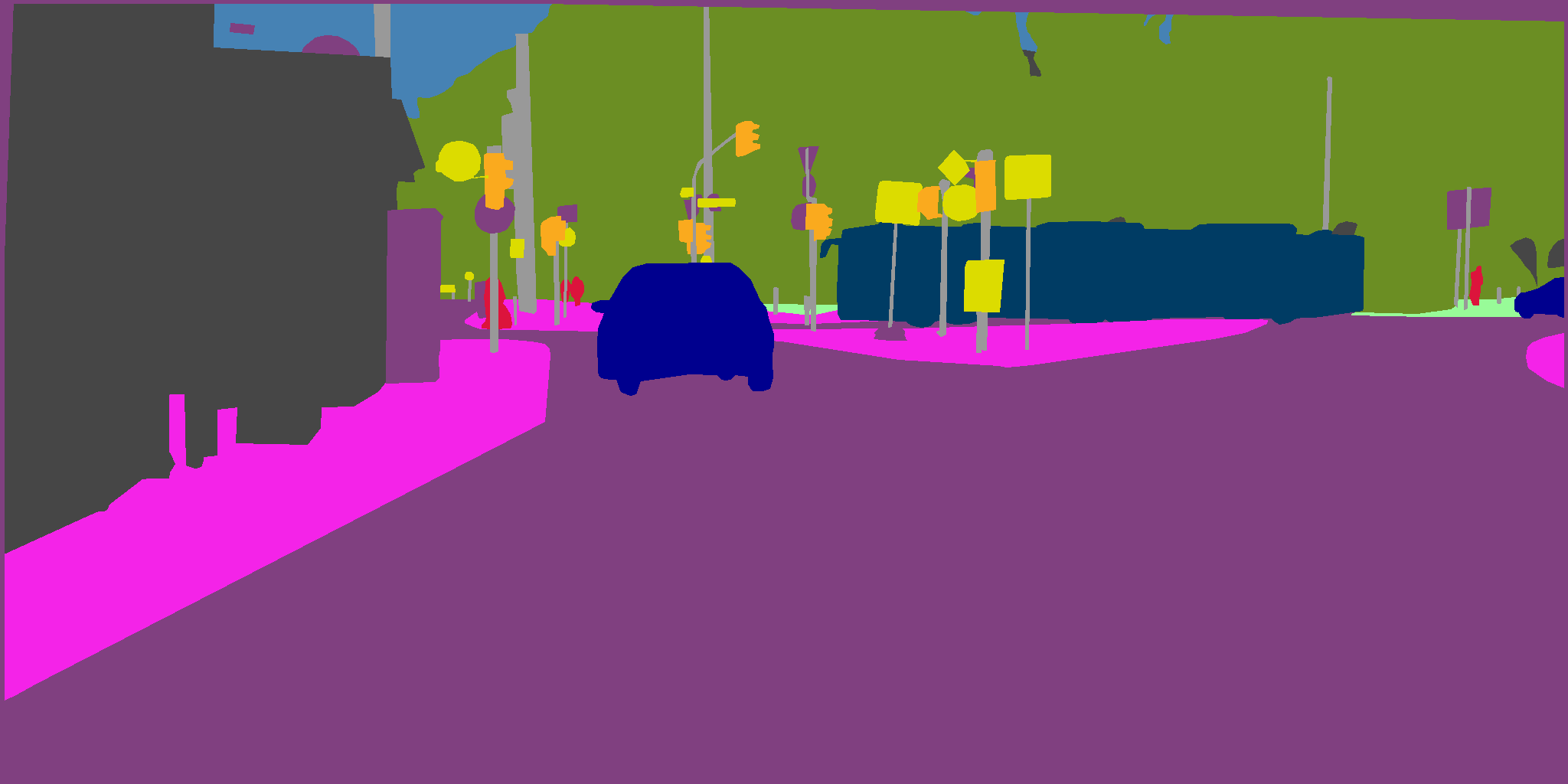} &
        \includegraphics[width=0.12\textwidth]{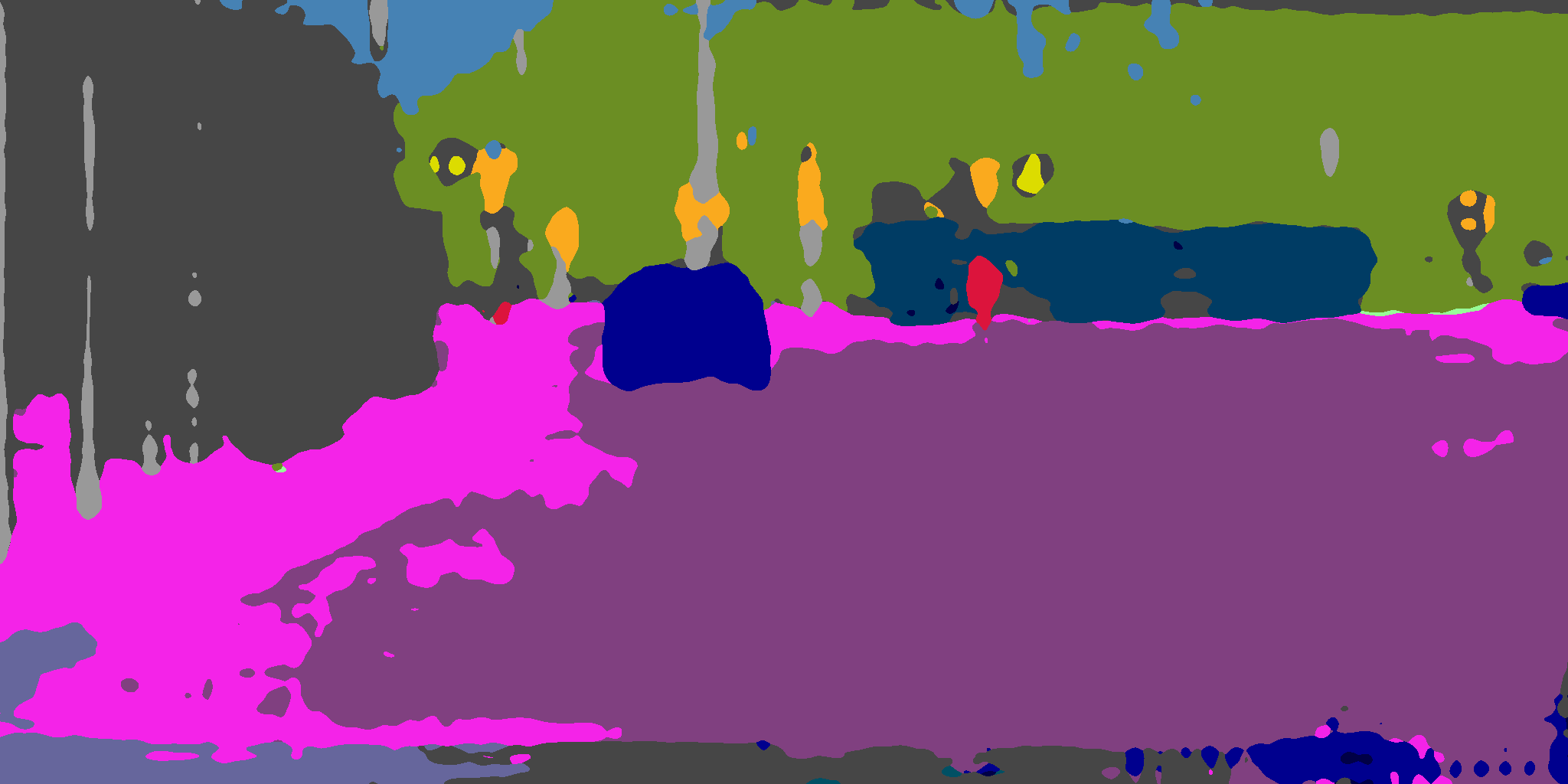} &
        \includegraphics[width=0.12\textwidth]{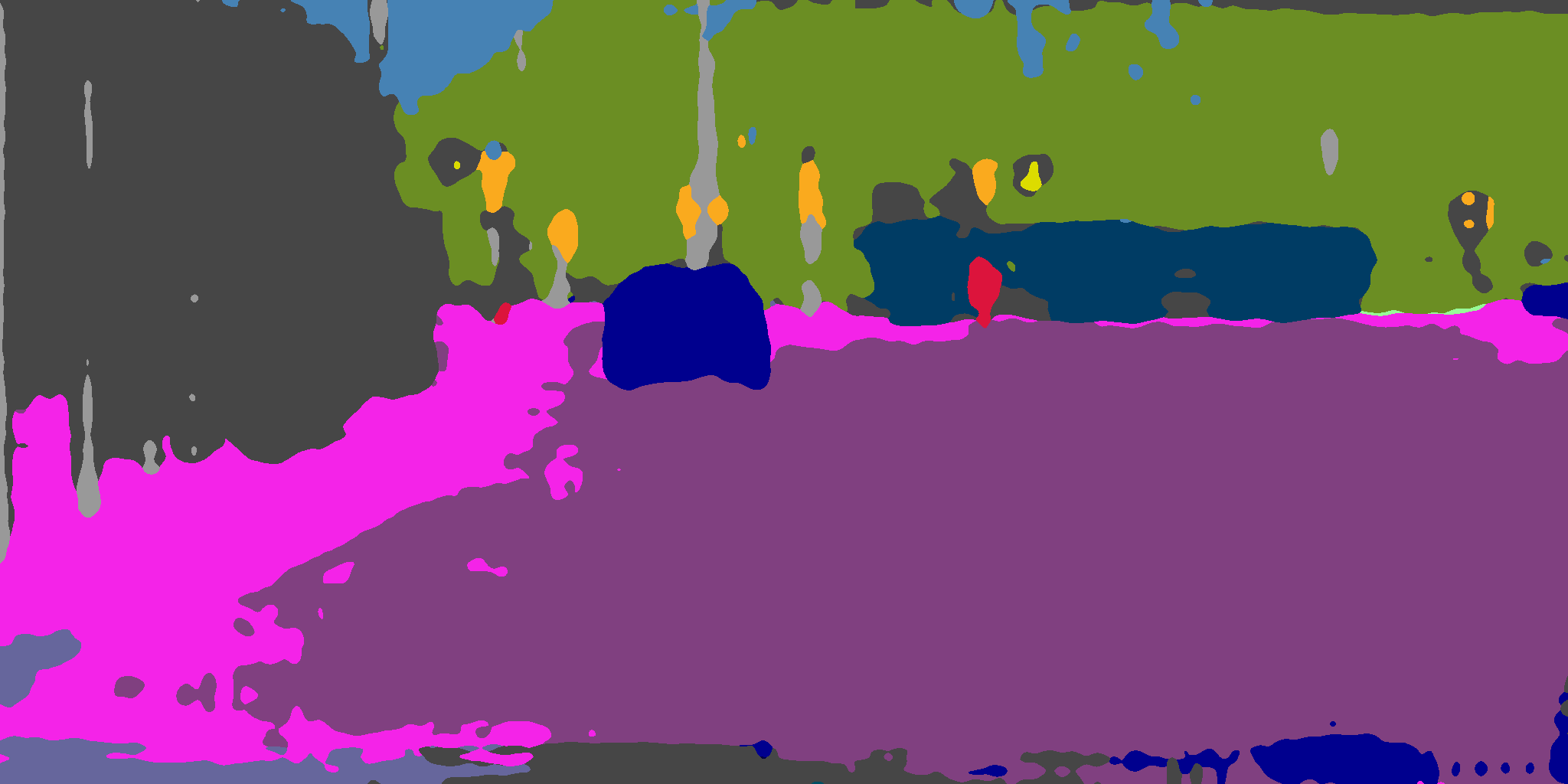} &
        \includegraphics[width=0.12\textwidth]{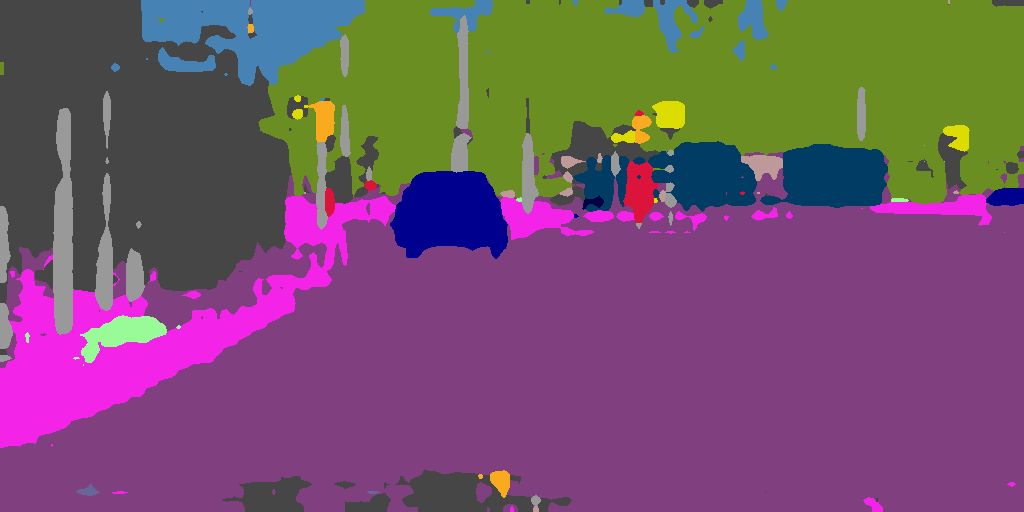} &
        \includegraphics[width=0.12\textwidth]{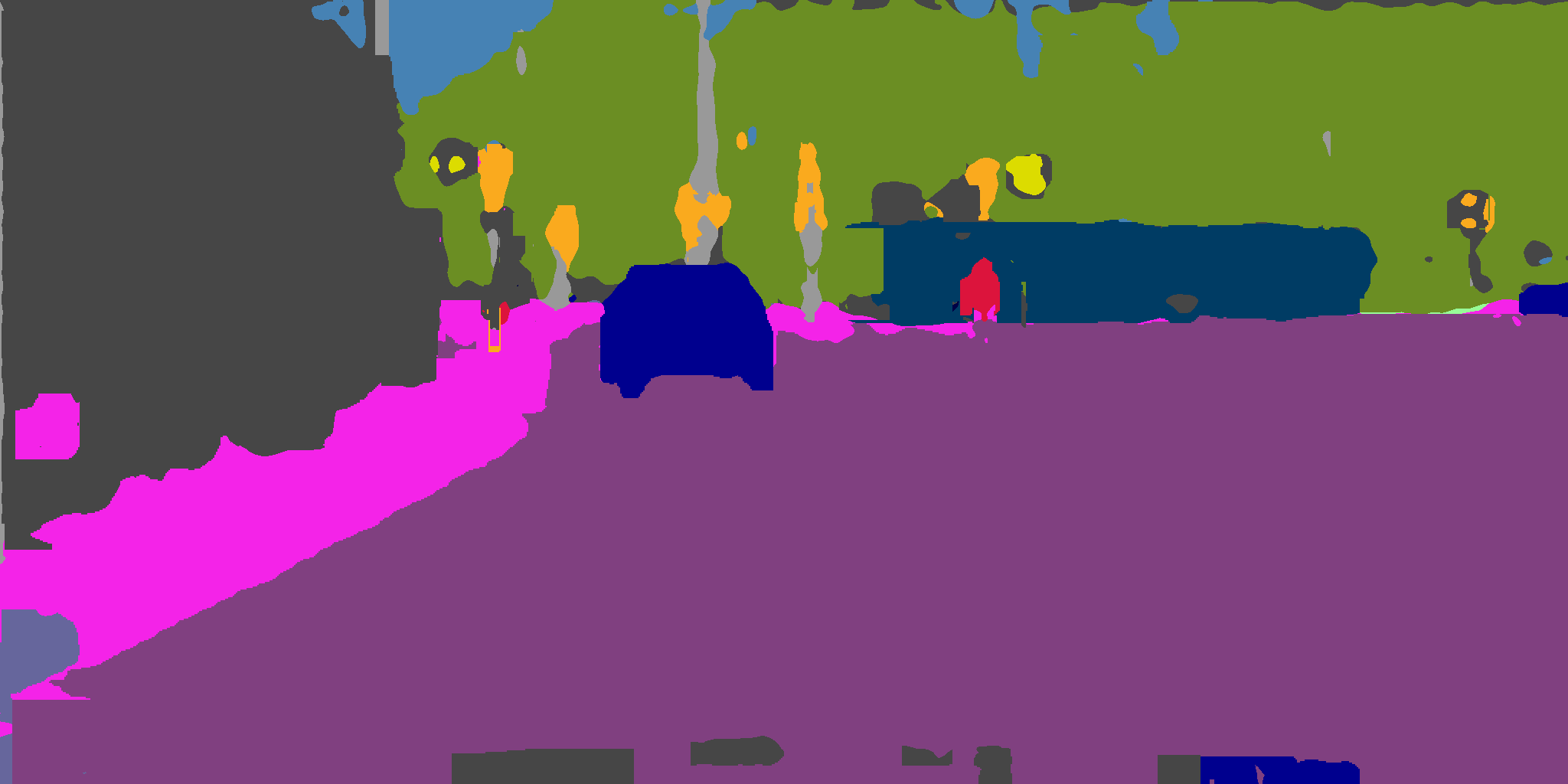} \\
        \raisebox{1\height}{\rotatebox{90}{\textbf{CS}}} &
        \includegraphics[width=0.12\textwidth]{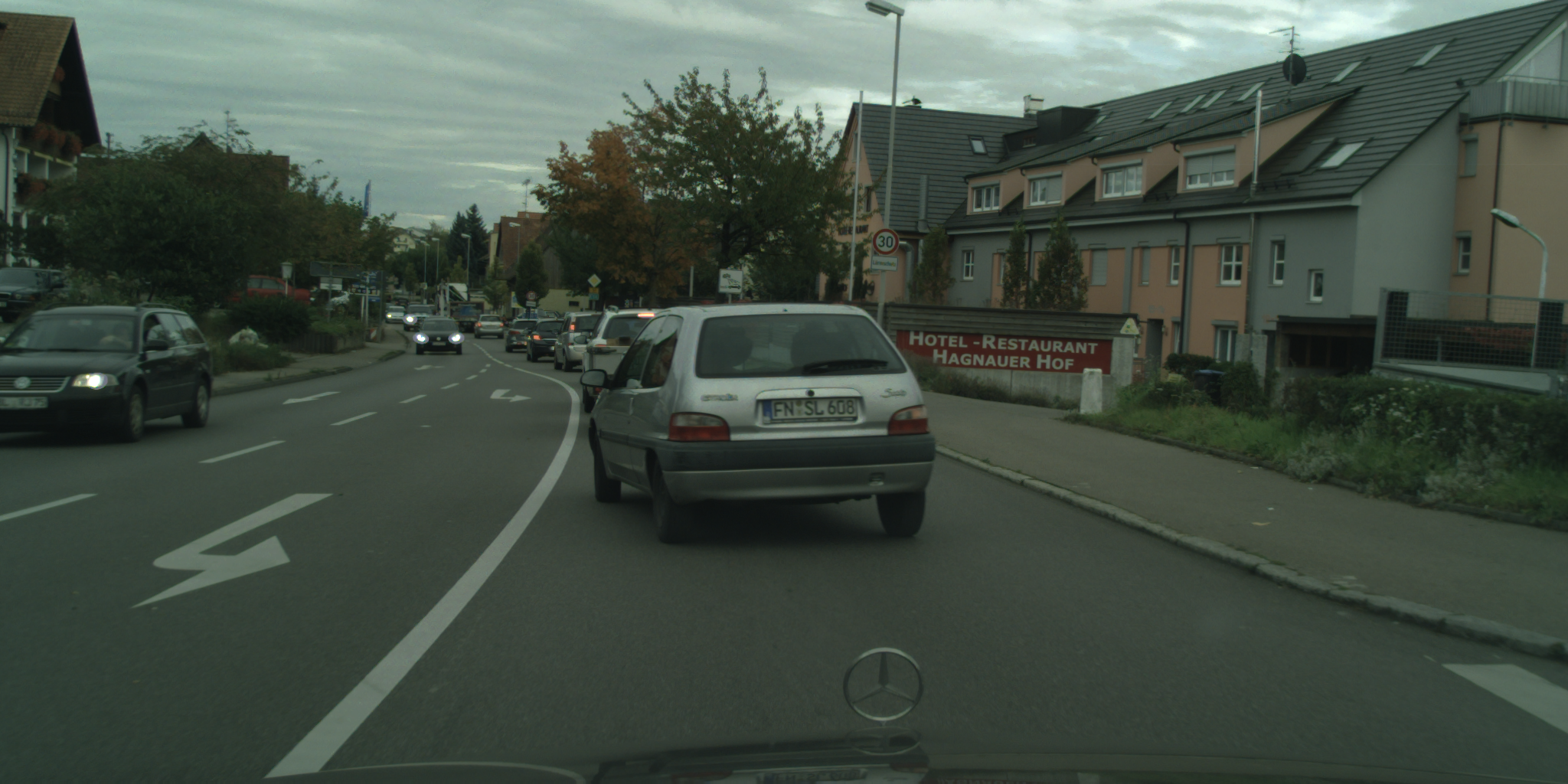} &
        \includegraphics[width=0.12\textwidth]{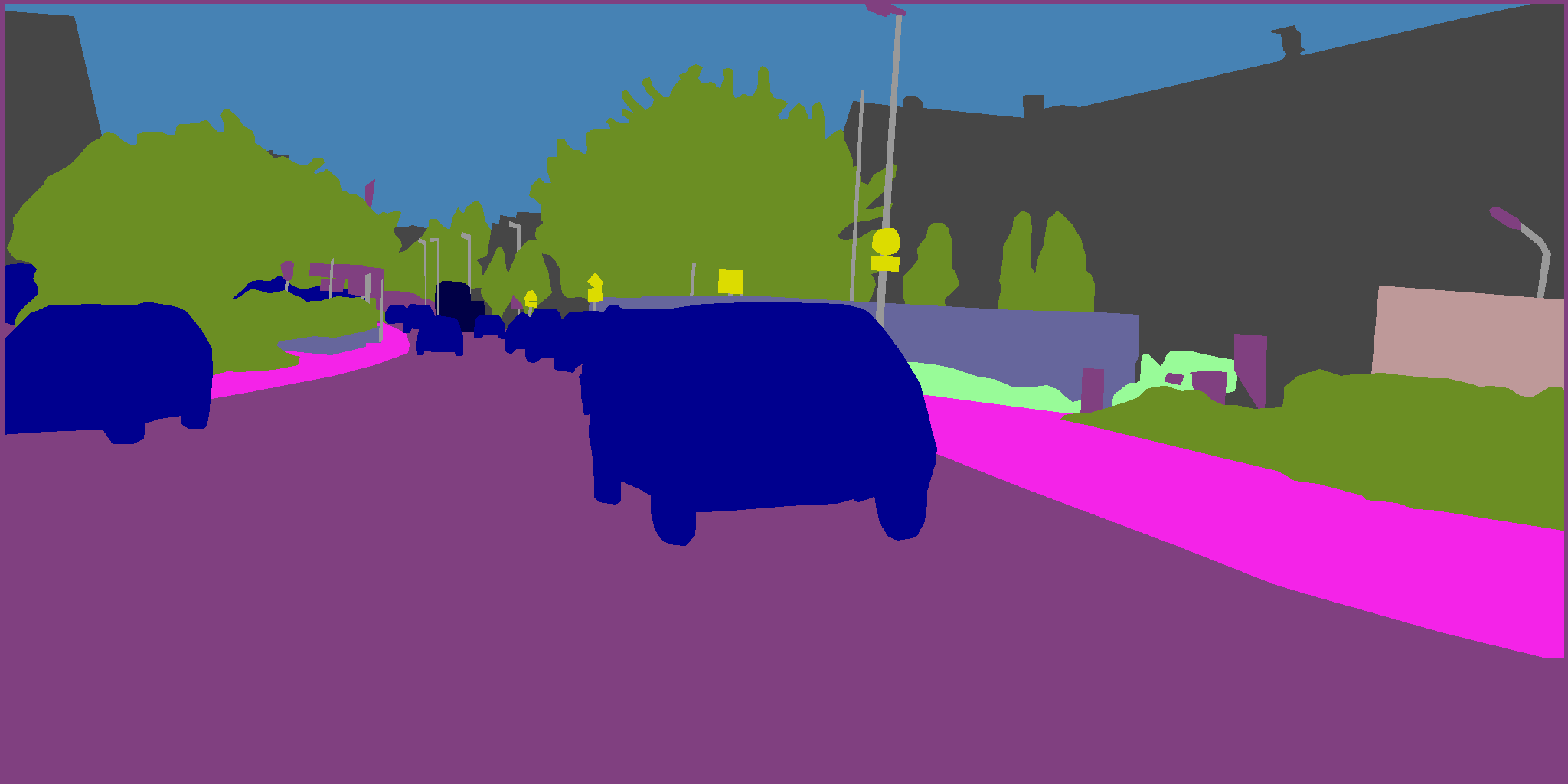} &
        \includegraphics[width=0.12\textwidth]{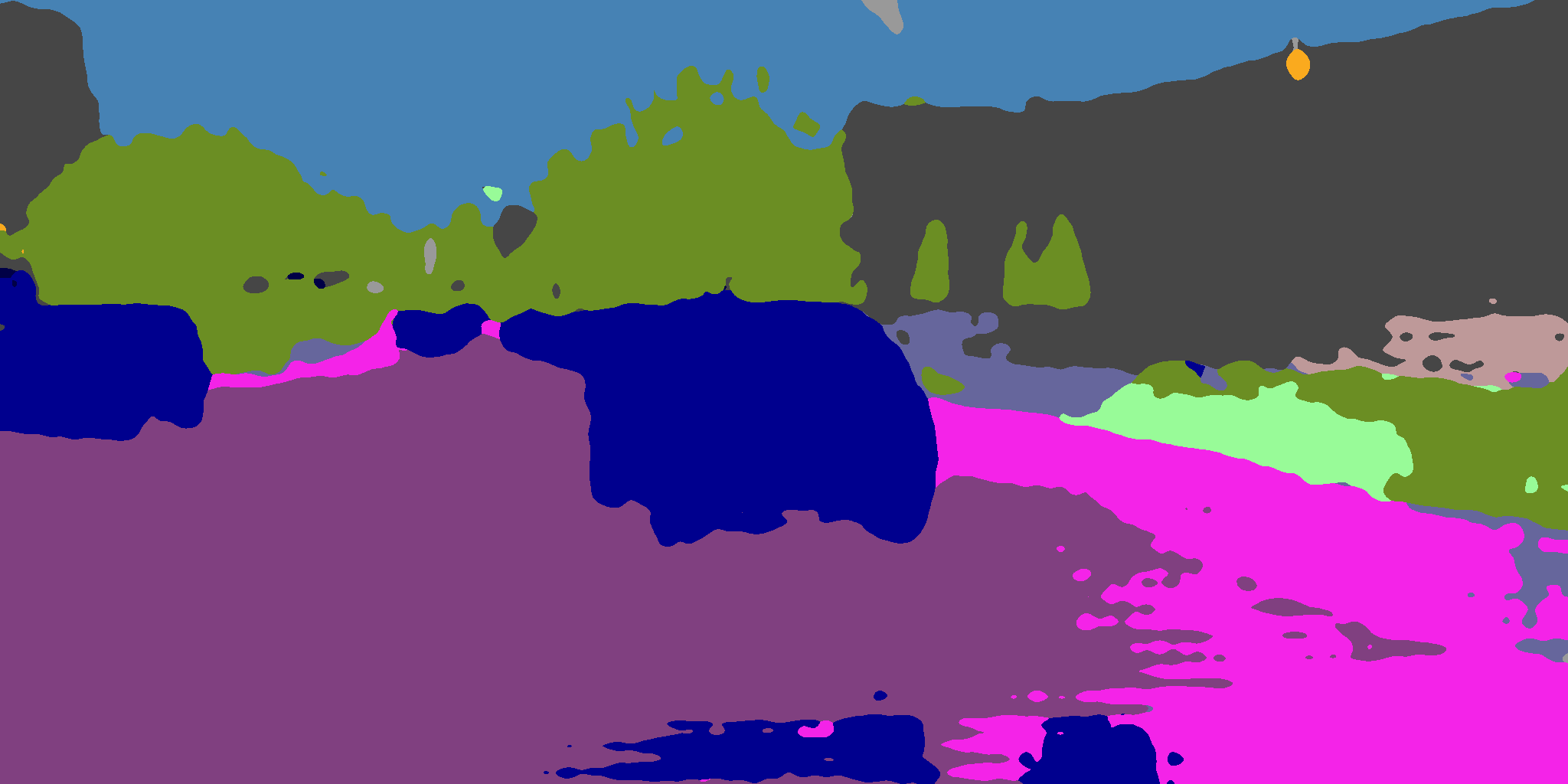} &
        \includegraphics[width=0.12\textwidth]{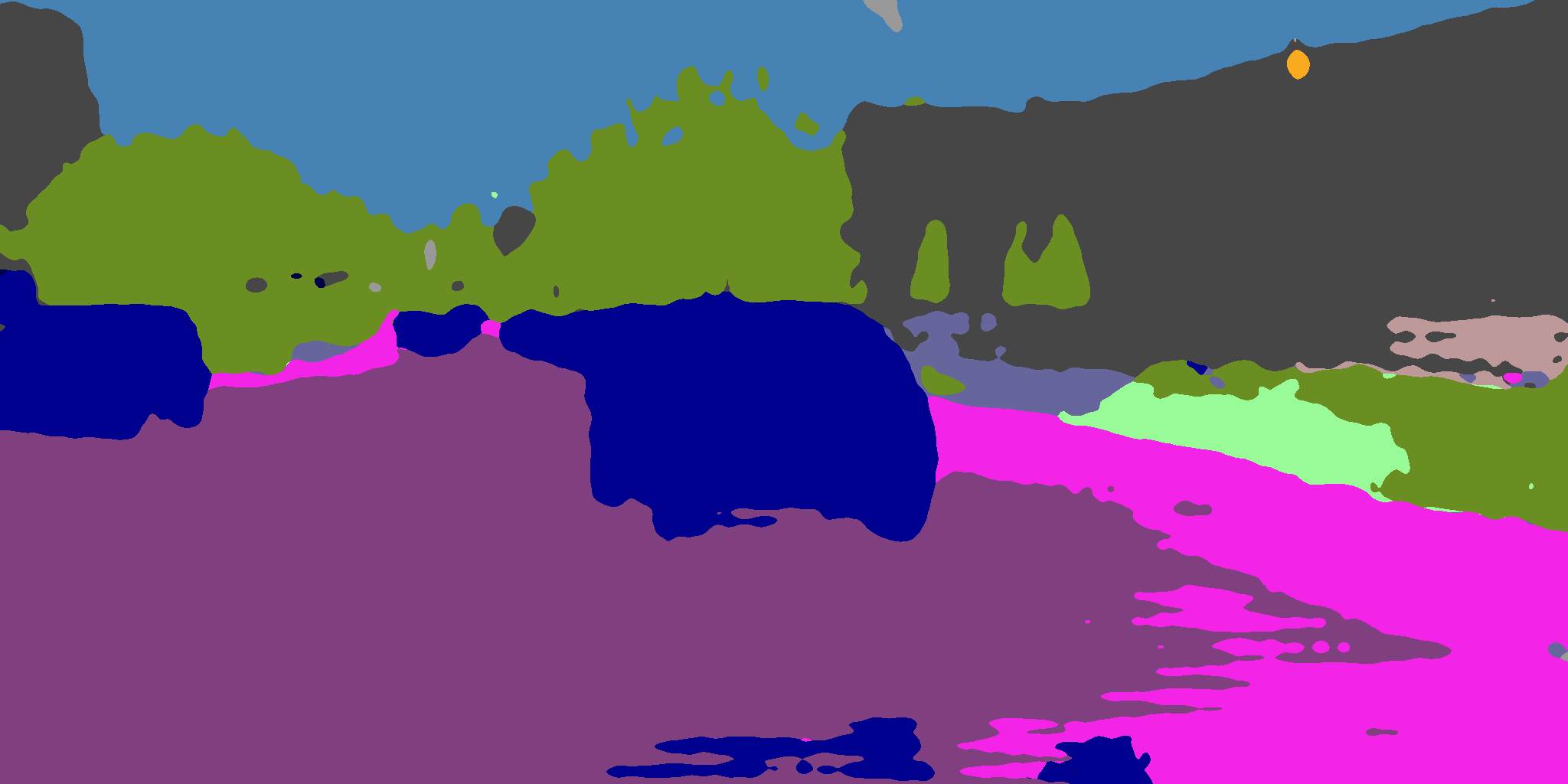} &
        \includegraphics[width=0.12\textwidth]{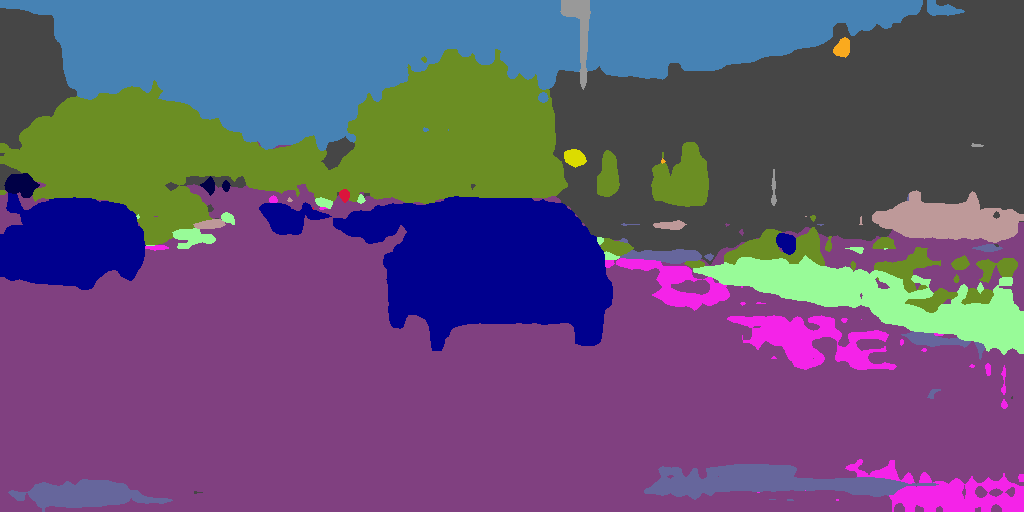} &
        \includegraphics[width=0.12\textwidth]{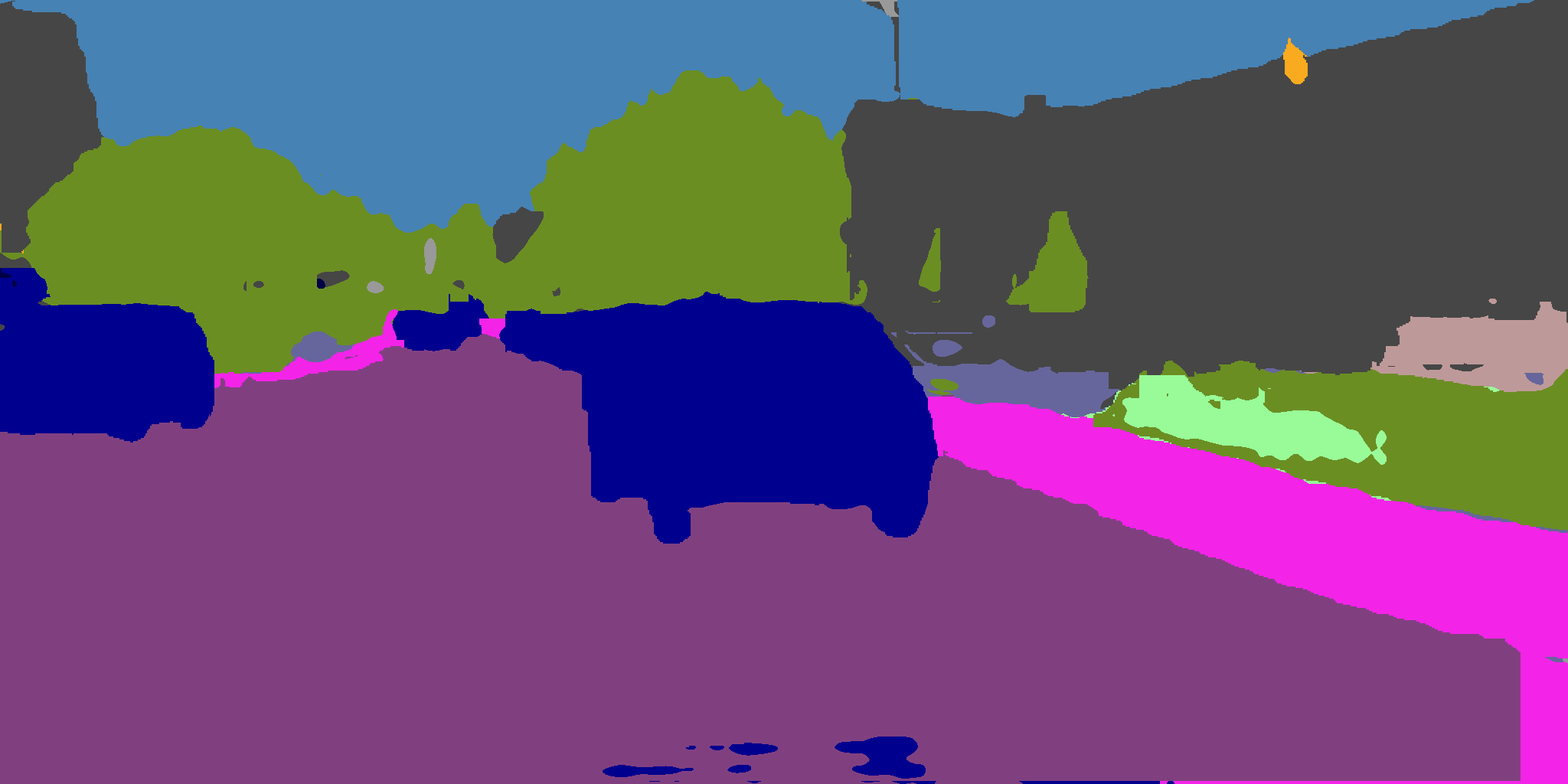} \\
        \raisebox{0.8\height}{\rotatebox{90}{\textbf{MV}}} &
       \includegraphics[width=0.12\textwidth]{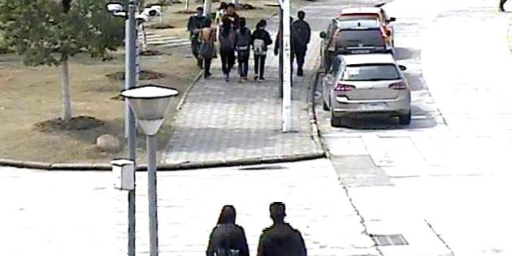} &
        \includegraphics[width=0.12\textwidth]{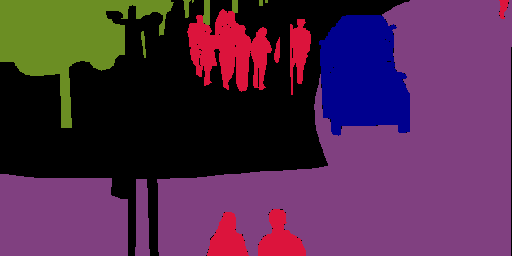} &
        \includegraphics[width=0.12\textwidth]{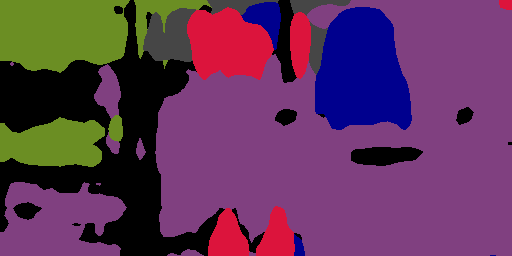} &
        \includegraphics[width=0.12\textwidth]{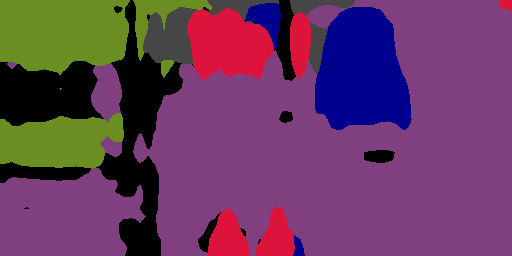} &
        \includegraphics[width=0.12\textwidth]{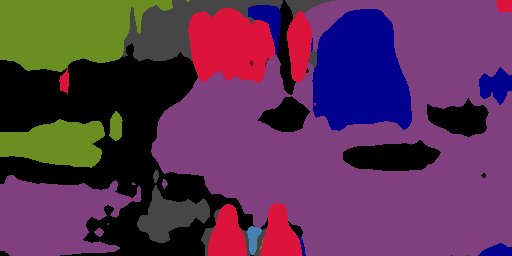} &
        \includegraphics[width=0.12\textwidth]{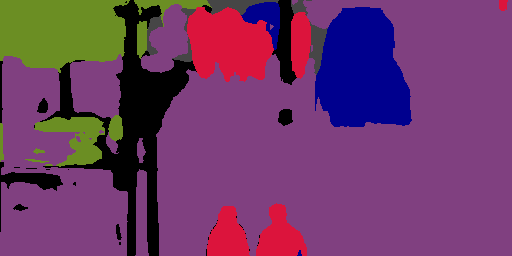} \\
        \raisebox{0.8\height}{\rotatebox{90}{\textbf{MV}}} &
        \includegraphics[width=0.12\textwidth]{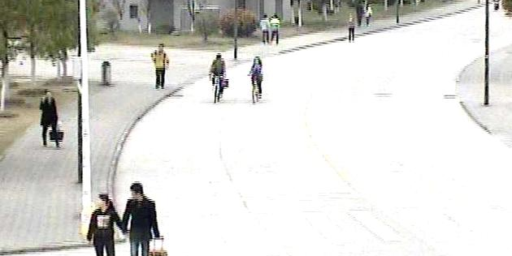} &
        \includegraphics[width=0.12\textwidth]{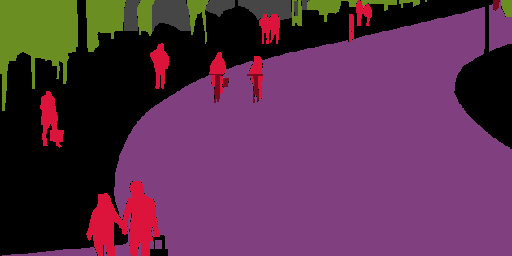} &
        \includegraphics[width=0.12\textwidth]{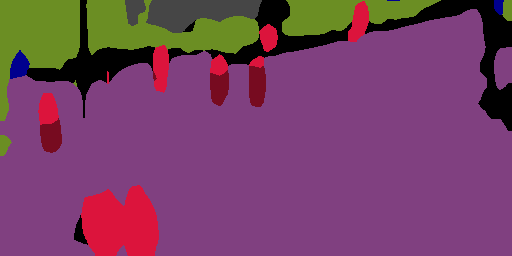} &
        \includegraphics[width=0.12\textwidth]{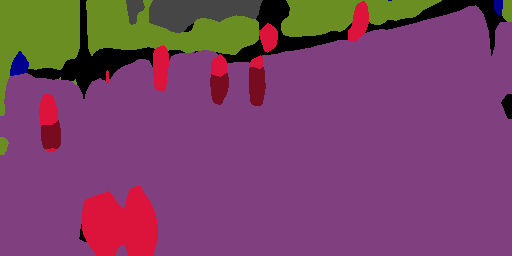} &
        \includegraphics[width=0.12\textwidth]{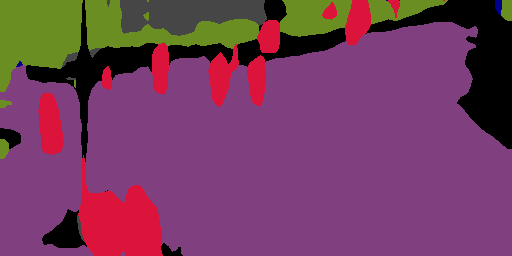} &
        \includegraphics[width=0.12\textwidth]{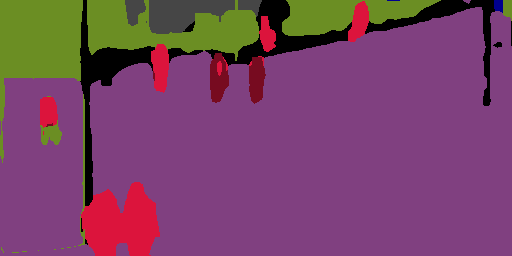}\\
    \end{tabular}
    \caption{\fontsize{9}{11}\selectfont Qualitative comparison on GTA-V → Cityscapes \cite{cordts2016cityscapes} and FMB \cite{liu2023multi} → MVSeg \cite{ji2023mvss} Dataset of TTDA methods.}
    \label{fig:qualitivegat5city}
\end{figure*}

\subsection{Integration with Other Frameworks}
Our method directly refines predictions, providing a simple, plug-and-play TTDA solution that requires neither backpropagation nor memory-banks. Despite its simplicity, it can seamlessly enhance existing TTDA, SFDA, or online-TTDA methods, both gradient- and non-gradient-based, boosting their performance.

Entropy-minimization methods like TENT \cite{wang2020tent} and EATA \cite{niu2022efficient} require multiple epochs to converge. In fast-paced applications, TestMate can refine the noisy, partially-adapted outputs from these methods, allowing the system to benefit from improved segmentation immediately, without waiting for full adaptation.

For pseudo-label-based methods, TestMate offers additional advantages. Many approaches \cite{song2023ecotta, wang2022continual, zhao2023towards} use a teacher–student setup, where the teacher generates pseudo-labels for the student to learn from. The teacher is periodically updated based on the student’s parameters. TestMate refines the teacher’s pseudo-labels, improving their quality, accelerating convergence, and boosting final accuracy.

\section{Experiments}

\subsection{Experimental Setup}

\subsubsection{Dataset}
Following prior works \cite{zhao2023towards, wang2023dynamically, chen2019domain}, we evaluate our method in a sim-to-real setting. We use the GTA-V dataset \cite{richter2016playing} as the source domain, consisting of 24,971 synthetic images annotated with 19 semantic classes. For the target domain, we utilize the Cityscapes dataset \cite{cordts2016cityscapes} which share the same 19 classes, allowing for direct cross-domain comparison. In addition to sim-to-real, we evaluate in a real-to-real setting using the FMB \cite{liu2023multi} and adapt to the dataset MVSeg dataset \cite{ji2023mvss}, a composition of multiple real-world datasets \cite{INO2012, hwang2015multispectral, li2019rgb}.

\subsubsection{Evaluation}
We assess performance using per-class Intersection over Union (IoU) and mean IoU (mIoU). TestMate is evaluated under three unsupervised settings. In Offline SFDA, adaptation is performed on the entire target dataset before evaluation. TTDA involves sequential adaptation using single-domain batches. Online TTDA addresses continuously shifting domains. For both TTDA and Online TTDA, performance metrics are computed for each batch and then averaged across all batches to reflect real-time, on-the-fly performance.

\subsubsection{Implementation Details}
For sim-to-real experiments, we adopt DeepLabV2 \cite{chen2017deeplab}, while real-to-real experiments employ DeepLabV3 \cite{chen2017rethinking}. Both models use a ResNet -101 \cite{he2016deep} backbone pretrained on ImageNet. For lightweight VFM within TestMate, we use FastSAM \cite{zhao2023fast} based on YOLOv8-seg \cite{jocher2023yolov8, 10533619} with an input resolution of 640, a confidence threshold of 0.1, and an IoU threshold of 0.6 for non-maximum suppression. Although additional parameters are reported in the ablation study, we primarily use the most efficient settings as required by the TTDA task. All training and evaluation are performed on an RTX 4090 GPU. For fairness, all methods are re-implemented with common source weights.

\subsection{Comparison with State of the Art}

\subsubsection{SFDA task}
For SFDA, we compare against strong baselines. URMDA \cite{fleuret2021uncertainty} extracts robust features through strong augmentations. SFDA \cite{liu2021source} transfers knowledge via a generator. SDF \cite{ye2021source} constructs virtual source information from high-confidence target samples. HCL \cite{huang2021model} applies historical contrastive learning on embeddings from current and past models. DS-ST \cite{zhao2023towards} dynamically updates a teacher–student framework using a copy–paste strategy \cite{ghiasi2021simple}.

\Cref{tab:miou-results_gta5-cityscapes} highlights the relative ease of the SFDA task compared to TTDA, as access to the full dataset and multiple training epochs significantly boosts performance. Among the methods, DS-DT \cite{zhao2023towards} achieves the highest accuracy, underscoring the value of dynamically updating the teacher model at the right moments from the student. While conceptually related to CoTTA \cite{wang2022continual}, which performs moderately in TTDA, DS-DT \cite{zhao2023towards} clearly outperforms it in this setting.

\begin{figure}[t]
  \centering
        \includegraphics[width=0.45\linewidth]{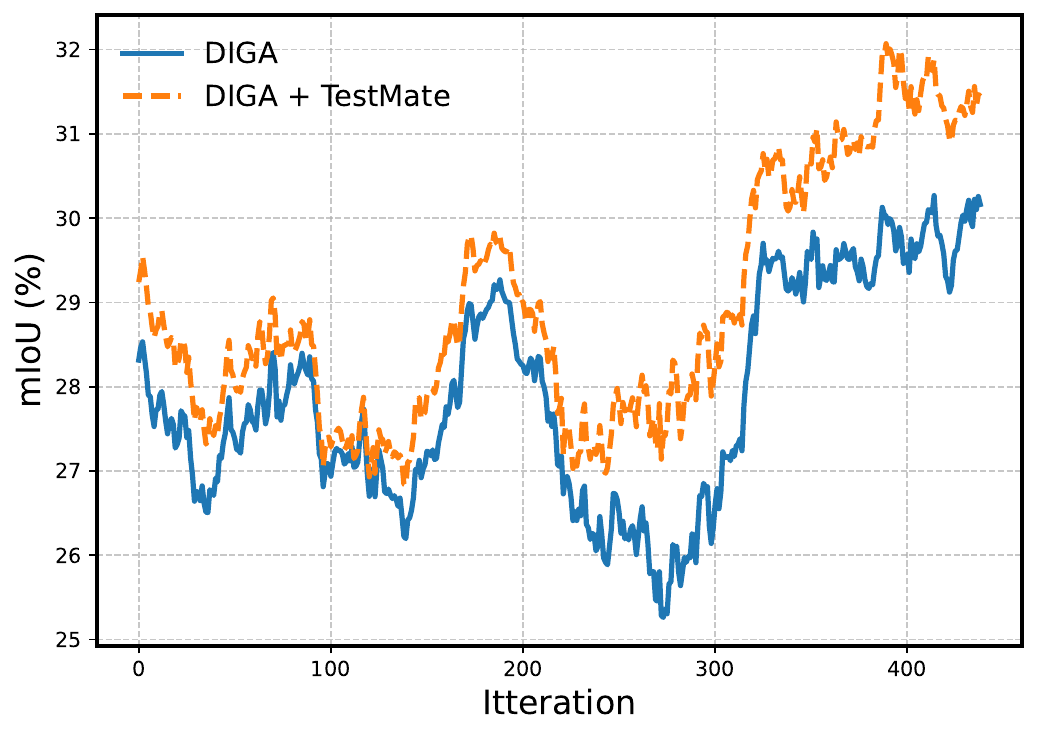} % half width in single-column
        \includegraphics[width=0.49\linewidth]{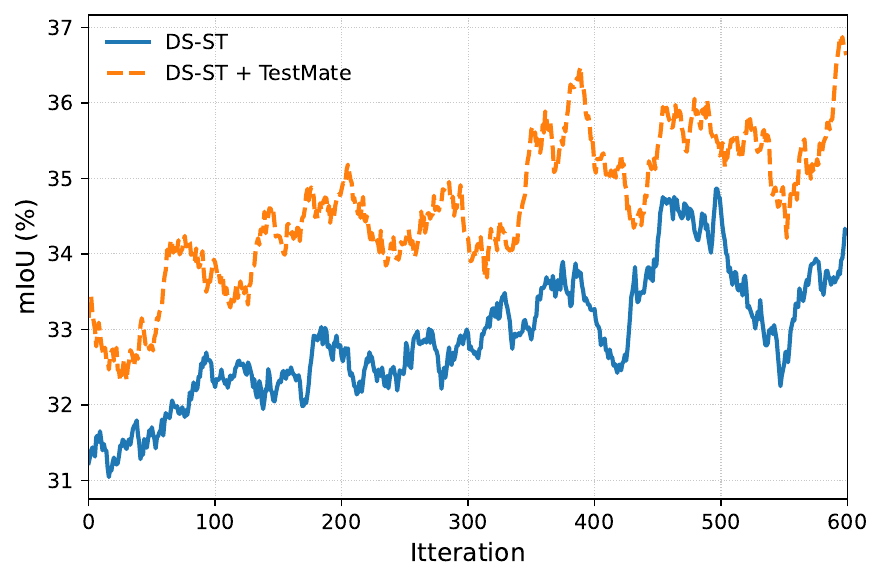} % half width in single-column
  \caption{\fontsize{9}{11}\selectfont Effect of TestMate (orange) when paired with (a) DIGA \cite{wang2023dynamically} (blue)  and (b) DS-DT \cite{zhao2023towards} (blue) in accuracy (mIoU) of intermediate produced pseudo-labels (GTA-V \cite{richter2016playing} → Cityscapes \cite{cordts2016cityscapes})}
  \label{fig:miou_comparisondsst}
\end{figure}

When combined with TestMate, the DS-DT \cite{zhao2023towards} + TestMate setup sets a new state-of-the-art, particularly improving results on medium and small classes. As shown in \cref{fig:miou_comparisondsst}, TestMate enhances the quality of pseudo-labels from the very first iteration, accelerating convergence and improving final accuracy.

\begin{table*}[ht!]
\centering
\footnotesize
\fontsize{9}{11}\selectfont % 9-point font for table content
\setlength{\tabcolsep}{2pt}
\begin{tabular}{l|c|ccccccccccc}
\toprule
Method & BP & background & vehicle & cycle & human & road & building & sky & vegetation & traffic light & Traffic Sign & mIoU \\
\midrule
Source & - & 12.8 & 59.3 & 6.7 & 36.1 & 52.4 & 45.3 & 55.5 & 52.3 & 9.0 & 6.1 & 33.6 \\
BSAdapt  & \xmark & 9.3 & 41.2 & 2.8 & 23.8 & 46.4 & 27.6 & 33.2 & 44.1 & 2.8 & 5.0 & 23.6  \\
TENT \cite{wang2020tent} & \checkmark & 9.7 & 61.5 & 5.3 & 36.7 & 51.6 & 45.5 & 55.8 & 52.2 & 6.8 & 5.7 & 33.1 \\
EATA \cite{niu2022efficient} & \checkmark & 12.4 & 59.5 & 6.5 & 36.3 & 52.3 & 45.4 & 55.5 & 52.4 & 8.8 & 6.3 & 33.6 \\
CoTTa \cite{wang2022continual} & \checkmark & 13 & 59.1 & 6.4 & 35.9 & 52.4 & 45.2 & 55.5 & 52.1 & 9.1 & 6.2 & 33.5  \\
DIGA \cite{wang2023dynamically} & \xmark & \textbf{16.6} & 61.1 & 5.2 & 34.8 & 53.2 & 46.1 & 56.3 & 53.2 & 1.4 & \textbf{10.6} & 33.9 \\
TestMate (Ours) & \xmark & 12.2 & 60.9 & \textbf{6.7} & 37.2 & 52.9 & 45.6 & 56.3 & 52.5 & \textbf{8.6} & 6.4 & 34.0 \\
TestMate (Ours) + CoTTA \cite{wang2022continual} & \checkmark &  11.3 & 62.6 & 5.7 & \textbf{38.1} & 52.8 & 45.4 & 56.3 & 51.9 & 6.6 & 7.0 & 33.8 \\
TestMate (Ours) + DIGA \cite{wang2023dynamically} & \xmark & 15.1 & \textbf{61.8} & 5.6 & 36.2 & \textbf{53.2} & \textbf{46.6} & \textbf{56.7} & \textbf{53.3} & 2.3 & 9.9 & \textbf{34.1} \\
\bottomrule
\end{tabular}
\caption{\fontsize{9}{11}\selectfont Experimental results for FMB \cite{liu2023multi} → MVSeg \cite{ji2023mvss} Dataset (shuffled test set) in Online TTDA settings with DeepLabV3 \cite{chen2017rethinking} (ResNet-101 \cite{he2016deep} backbone). BP denotes if the method utilizes backpropagation to achieve adaptation}
\label{tab:miou-results-onttdashuf}
\end{table*}

\begin{table}[t]
\begin{center}
\renewcommand{\arraystretch}{1.2}
\fontsize{9}{11}\selectfont % 9-point font for table content
\resizebox{\columnwidth}{!}{
\begin{tabular}{l|ccc|c}
\hline
 & \multicolumn{3}{c|}{$\texttt{t} \xrightarrow{\hspace{3.9cm}}$} & \\
\cline{2-4}
Method & INO & KAIST & RGBT234 & Mean \\
\hline
Source & 27.9 & 34.1 & 32.3 & 31.4 \\
TENT \cite{wang2020tent} & 29.5 & 35.6 & 31.9 & 32.3 \\
EATA \cite{niu2022efficient} & 27.9 & 36.2 & 32.2 & 32.1 \\
CoTTa \cite{wang2022continual} & 27.5 & 35.9 & 32.1 & 31.8 \\
DIGA \cite{wang2023dynamically} & \textbf{32.6} & 33.6 & 31.6 & 32.6 \\
TestMate (Ours) & 29.0 & 35.0 & \textbf{33.8} & 32.6 \\
TestMate + CoTTA \cite{wang2022continual} & 28.4 & 34.8 & 33.6 & 32.3 \\
TestMate + DIGA \cite{wang2023dynamically} & 32.0 & \textbf{36.8} & 32.3 & \textbf{33.7} \\
\hline
\end{tabular}
}
\end{center}
\caption{\fontsize{9}{11}\selectfont
Performance (mIoU \%) of online TTDA methods on the MVSeg \cite{ji2023mvss} dataset, where frames from INO \cite{INO2012}, KAIST \cite{hwang2015multispectral}, and RGBT234 \cite{li2019rgb} arrive sequentially over time ($\texttt{t}$). Each entry shows results after adaptation to each dataset and the overall mean.
}
\label{tab:miou-comparison}
\end{table}

\subsubsection{TTDA task}
For TTDA, we compare against BSAdapt, a simple baseline that updates batch statistics during evaluation and works well on classification tasks. We also include the TENT \cite{wang2020tent} method, which uses entropy minimization, and EATA \cite{niu2022efficient}, which filters out noisy pixels before minimization. The CoTTA \cite{wang2022continual} model incorporates consistency between student and teacher networks by generating pseudolabels from augmented target inputs. For non-gradient-based methods, we compare with DIGA  \cite{wang2023dynamically} , which uses a prototype memory-bank built from target inputs. We present our method on its own, and also integrate it with both a gradient-based approach (CoTTA \cite{wang2022continual}) and a non-gradient-based approach (DIGA) \cite{wang2023dynamically}.

From \cref{tab:miou-results_gta5-cityscapes}, we observe that methods well established for classification tasks, such as TENT \cite{wang2020tent}, EATA \cite{niu2022efficient}, and CoTTA \cite{wang2022continual}, offer only marginal improvements over the source (unadapted) model when applied to semantic segmentation. This highlights the greater challenge and relative underexploration of TTDA for segmentation.

\begin{figure}[t]
    \centering
    % First row
    \begin{subfigure}[t]{0.096\textwidth}
        \includegraphics[width=\linewidth]{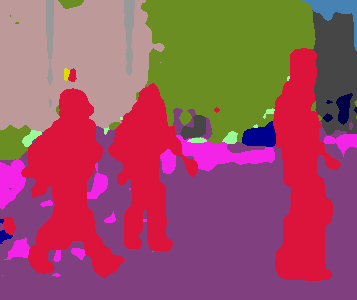}
        \caption{}
        \label{fig:sub1}
    \end{subfigure}
    \begin{subfigure}[t]{0.1\textwidth}
        \includegraphics[width=\linewidth]{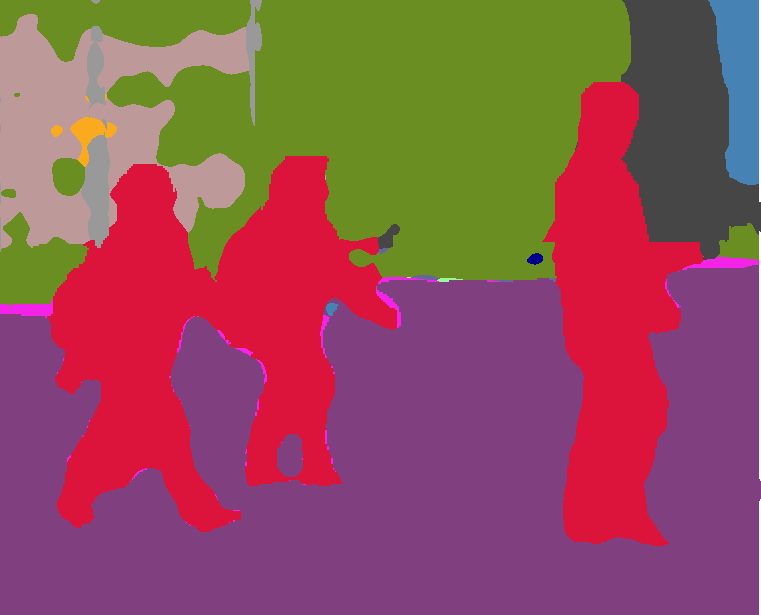}
        \caption{}
        \label{fig:sub2}
    \end{subfigure}
    % Second row
    \begin{subfigure}[t]{0.112\textwidth}
        \includegraphics[width=\linewidth]{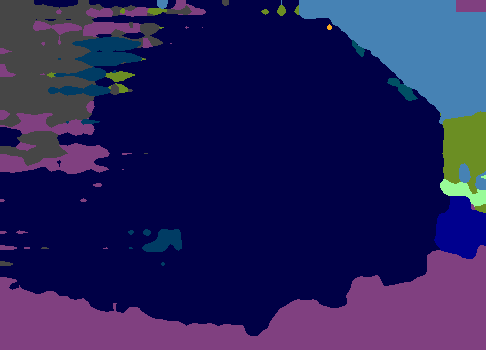}
        \caption{}
        \label{fig:sub3}
    \end{subfigure}
    \begin{subfigure}[t]{0.113\textwidth}
        \includegraphics[width=\linewidth]{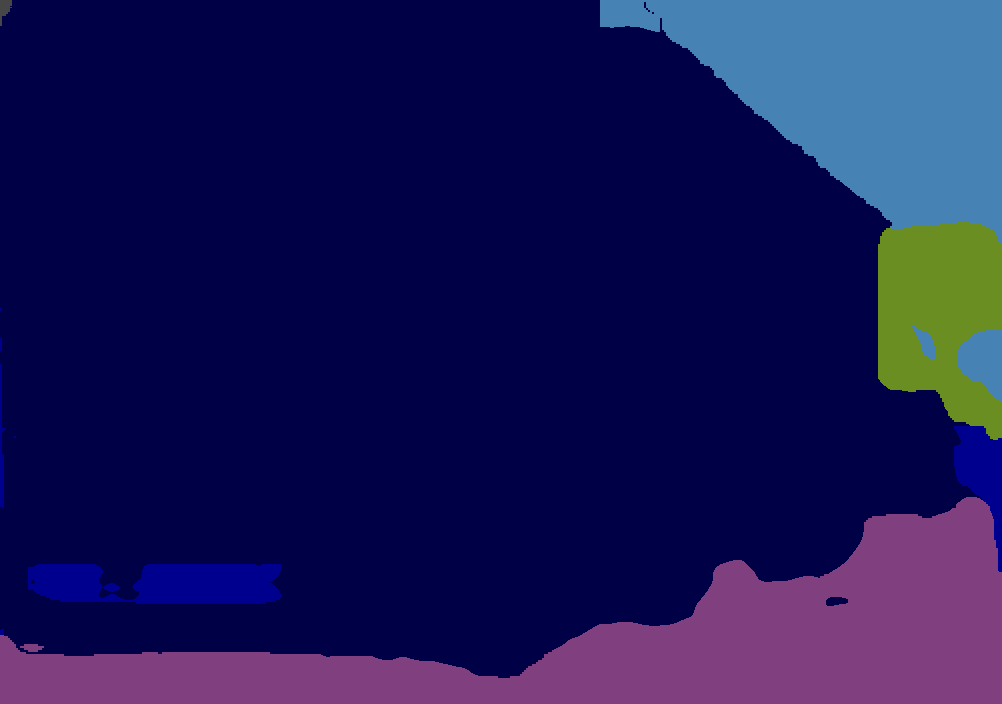}
        \caption{}
        \label{fig:sub4}
    \end{subfigure}

    \caption{\fontsize{9}{11}\selectfont Visual example of TestMate (b, d) boosting boundaries and increasing coherence of semantic instances in the Cityscapes dataset compared to DIGA \cite{wang2023dynamically} method (a, c).}
    \label{fig:digatestmatehuman}
\end{figure}

Among the baseline approaches, DIGA \cite{wang2023dynamically} stands out, outperforming others by 4-5\%, highlighting that gradient-free methods can provide more stable and robust adaptation. Our method is not only gradient-free but also does not rely on a memory bank, yet still achieves nearly a 3\% improvement over the unadapted model and surpasses gradient-based approaches by over 2\%. While performance on large, abstract classes such as road is slightly lower than DIGA \cite{wang2023dynamically}, we observe substantial gains on smaller, well-defined classes like person and rider. These improvements are further illustrated in \cref{fig:digatestmatehuman}, which demonstrates the effectiveness of VFM-derived masks in refining object boundaries. As shown in the first row of \cref{fig:qualitivegat5city}, TestMate exhibits higher spatial coherence in segmenting medium-sized classes, whereas DIGA \cite{wang2023dynamically} tends to fragment them into smaller regions.

When combined with the CoTTA \cite{wang2022continual} technique, by refining pseudo-labels used to train the student model, our method further improves accuracy beyond standard CoTTA \cite{wang2022continual}, although the margin is relatively modest compared to our standalone approach. Finally, when integrated with DIGA\cite{wang2023dynamically}, our method yields an additional 0.9\% mIoU gain, setting a new state-of-the-art. This improvement is evident across both large categories, like road and smaller ones, such as person, car, and rider.

As shown in \cref{tab:iteration_time}, TestMate adds minimal computational overhead due to its lightweight VFM, which runs in parallel with pretrained model inference. 

\begin{table}[h]
\centering

\fontsize{9}{11}\selectfont % 9-point Roman (usually Times New Roman/Latin Modern) for table content
\begin{tabular}{lc}
\toprule
Method & Time / Iter. (ms) \\
\midrule
TENT \cite{wang2020tent} & 310 \\
CoTTA \cite{wang2022continual} & 400 \\
DIGA \cite{wang2023dynamically} & 55 \\
TestMate + DIGA \cite{wang2023dynamically} & 60 \\
\bottomrule
\end{tabular}
\caption{\fontsize{9}{11}\selectfont Per-sample computation time (ms) for TTDA methods on RTX4090.}
\label{tab:iteration_time}
\end{table}

While TestMate can be applied iteratively to enhance refinement, \cref{tab:number_iter} shows that most of the gains occur in the first iteration. Unlike other iterative methods such as TENT\cite{wang2020tent}, TestMate remains stable across many iterations. Its performance does not degrade even after numerous steps, ensuring that smaller objects are not suppressed by larger, more confident ones with more iterations.

\begin{table}[ht!]
\centering
\fontsize{9}{11}\selectfont % 9-point font for table content
\resizebox{\columnwidth}{!}{
\begin{tabular}{lcccccc}
\toprule
Method & 1 Iter & 3 Iter & 5 Iter & 7 Iter & 10 Iter & 20 Iter \\
\midrule
TENT \cite{wang2020tent}      & 36.76 & 36.80 & 37.00 & 37.10 & 37.20 & 36.90 \\
TestMate  & 39.45 & 39.50 & 39.65 & 39.65 & 39.64 & 39.63 \\
\bottomrule
\end{tabular}
}
\caption{\fontsize{9}{11}\selectfont
Performance mIoU (\%) of TENT \cite{wang2020tent} and TestMate algorithms as a function of refinement iterations per sample on TTDA task GTA-V \cite{richter2016playing} → Cityscapes \cite{cordts2016cityscapes}.
}
\label{tab:number_iter}
\end{table}

\subsubsection{Online TTDA task}
The final evaluation setting for TestMate is online-TTDA. In this task, a DeepLabV3 \cite{chen2017rethinking} model is trained on the FMB dataset \cite{liu2023multi} and evaluated on MVSeg, a benchmark comprising three datasets, each representing a distinct distribution. This setup assesses the robustness of adaptation methods under continuous distribution shift. We consider two subcases: (i) the target data is randomly shuffled, and (ii) the target distributions arrive sequentially. The methods considered are similar to the previous case, as most are designed to counter catastrophic forgetting and remain effective under distribution shifts.

The shuffled setting represents an extreme case of continual target distribution, requiring adaptation to multiple representations of target classes. In this scenario, from \cref{tab:miou-results-onttdashuf} we see that CoTTA \cite{wang2022continual}, TENT \cite{wang2020tent}, and EATA \cite{niu2022efficient} fail to adapt, performing worse than the source model. DIGA \cite{wang2023dynamically} remains competitive but lags slightly behind naive TestMate, while their combination achieves the best results. Unlike the previous TTDA task, TestMate surpasses DIGA \cite{wang2023dynamically} with memory-bank, highlighting its key strength: instance-based adaptation that remains robust under distribution shifts.

In the second online-TTDA setting, data from INO \cite{INO2012}, KAIST \cite{hwang2015multispectral}, and RGBT234 \cite{li2019rgb}  arrive sequentially. Here, TENT, EATA, and CoTTA outperform the source model, indicating their ability to cope with slower distribution shifts. DIGA\cite{wang2023dynamically} achieves the highest mIoU on the first dataset, but its performance drops sharply as the distribution changes, suggesting that its memory-bank limits fast adaptation. Naive TestMate lags behind DIGA \cite{wang2023dynamically} by 3.6\% mIoU on the first dataset but maintains consistent improvements over the source model in later datasets, achieving the best results on the final one and matching DIGA’s \cite{wang2023dynamically} mean mIoU. Combining DIGA \cite{wang2023dynamically} with TestMate once again yields the highest performance, surpassing both by 1.1\% mIoU.

\subsection{Ablation Study}
In \cref{tab:algoablat}, we present an ablation study evaluating the impact of key algorithmic design choices: mask ordering, refinement strategy, and entropy filtering. Among these, mask ordering has the most significant effect. Specifically, prioritizing smaller masks to refine the model’s segmentation output reduces the risk of a single mask from the VFM capturing multiple objects simultaneously, leading to cleaner segmentation ouptut. Both the refinement strategy and entropy filtering contribute comparably, each improving the final accuracy of the TestMate algorithm. Notably, these two components seem to work in complementary ways: each one individually improves mIoU by about +1\%, and using them together provides an additional, though smaller, performance boost.
\begin{table}[ht!]
\centering
\fontsize{9}{11}\selectfont
\resizebox{\columnwidth}{!}{
\begin{tabular}{|c|c|c|c|}
\hline
Order & Refinement & Entropy Filtering & mIoU \\
\hline
Descending & Hard & \ding{55} &  37.0 \\
Descending & Hard & \ding{51} &  39.0\\
Descending & Soft & \ding{55} &  38.8\\
Descending & Soft & \ding{51} &  39.2 \\
Ascending & Hard & \ding{55} &  38.1 \\
Ascending & Hard & \ding{51} &  39.4 \\
Ascending & Soft & \ding{55} &  39.3 \\
Ascending & Soft & \ding{51} &  \textbf{39.6} \\
\hline
\end{tabular}
}
\caption{\fontsize{9}{11}\selectfont Ablation study on TestMate components. We report mIoU on the TTDA task (GTA-V \cite{richter2016playing}~$\rightarrow$~Cityscapes \cite{cordts2016cityscapes}).}
\label{tab:algoablat}
\end{table}

In \cref{tab:vfm_ablation}, we present an ablation study on the impact of VFM model size. We compare two versions of FastSAM's \cite{zhao2023fast} backbone: small (YOLOv8-s) and medium (YOLOv8-x). Besides model size, input resolution also affects inference speed and segmentation accuracy. For the small model, the highest mIoU is achieved at an input resolution of 640, whereas increasing the resolution beyond this decreases segmentation performance. Conversely, for the medium model, the best mIoU is obtained at 1280 resolution, but with a significant reduction in inference speed.

Based on these findings, we use YOLOv8-s (640 px) for the main experiments, as it balances mIoU, speed, and model size. Fast inference is essential to be comparable to or faster than the main model to enable parallelization without degrading overall TTDA performance.

\begin{table}[ht!]
\centering
\fontsize{9}{11}\selectfont
\resizebox{\columnwidth}{!}{
\begin{tabular}{lcccc}
\toprule
Model & Input Res. (px) & mIoU (\%) & Speed (FPS) & Params (M) \\
\midrule
YOLOv8-s & 320  & 38.4 & 100 & 11\\
YOLOv8-s & 640  & 39.6 & 85 & 11\\
YOLOv8-s & 1280 & 39.0 & 45 & 11\\
YOLOv8-x & 320  & 38.8 & 84 & 68\\
YOLOv8-x & 640  & 40.0 & 54 & 68\\
YOLOv8-x & 1280 & 40.5 & 17 & 68\\
\bottomrule
\end{tabular}
}
\caption{\fontsize{9}{11}\selectfont
Ablation study of lightweight VFM size and input resolution: mIoU on the TTDA task (GTA-V \cite{richter2016playing}~$\rightarrow$~Cityscapes\cite{cordts2016cityscapes}) and inference speed (FPS) for YOLOv8 variants at different input resolutions.
}
\label{tab:vfm_ablation}
\end{table}

\section{Conclusion}
In this work, we introduced TestMate, a novel and effective algorithm that leverages lightweight Vision Foundation Models (VFMs) for the Test-Time Domain Adaptation (TTDA) task. TestMate utilizes high-quality, unlabeled semantic instances detected by the VFM and intelligently fuses this information with the unadapted model outputs, refining segmentation masks that are otherwise distorted by domain shifts. It operates as a gradient-free, instance-based method, improving segmentation accuracy on the fly, and can be seamlessly integrated with existing TTDA methods to further boost performance. TestMate achieved new state-of-the-art results across three challenging tasks: Source-Free Domain Adaptation (SFDA), TTDA, and online-TTDA. Future work could explore stronger VFMs, their integration with LLMs for TTDA, and alternative strategies for fusing VFM outputs with segmentation networks.
{
    \small
    \bibliographystyle{ieeenat_fullname}
    \bibliography{main}

@String(CVPR= {IEEE Conf. Comput. Vis. Pattern Recog.})

@String(NIPS= {Adv. Neural Inform. Process. Syst.})

@String(CVPR  = {CVPR})

@String(NIPS  = {NeurIPS})

@article{kouw2019review,
  title={A review of domain adaptation without target labels},
  author={Kouw, Wouter M and Loog, Marco},
  journal={IEEE transactions on pattern analysis and machine intelligence},
  volume={43},
  number={3},
  pages={766--785},
  year={2019},
  publisher={IEEE}
}

@article{li2024comprehensive,
  title={A comprehensive survey on source-free domain adaptation},
  author={Li, Jingjing and Yu, Zhiqi and Du, Zhekai and Zhu, Lei and Shen, Heng Tao},
  journal={IEEE Transactions on Pattern Analysis and Machine Intelligence},
  year={2024},
  publisher={IEEE}
}

@inproceedings{gao2023back,
  title={Back to the source: Diffusion-driven adaptation to test-time corruption},
  author={Gao, Jin and Zhang, Jialing and Liu, Xihui and Darrell, Trevor and Shelhamer, Evan and Wang, Dequan},
  booktitle={Proceedings of the IEEE/CVF Conference on Computer Vision and Pattern Recognition},
  pages={11786--11796},
  year={2023}
}

@inproceedings{tsai2024gda,
  title={GDA: Generalized Diffusion for Robust Test-time Adaptation},
  author={Tsai, Yun-Yun and Chen, Fu-Chen and Chen, Albert YC and Yang, Junfeng and Su, Che-Chun and Sun, Min and Kuo, Cheng-Hao},
  booktitle={Proceedings of the IEEE/CVF Conference on Computer Vision and Pattern Recognition},
  pages={23242--23251},
  year={2024}
}

@inproceedings{wang2022continual,
  title={Continual test-time domain adaptation},
  author={Wang, Qin and Fink, Olga and Van Gool, Luc and Dai, Dengxin},
  booktitle={Proceedings of the IEEE/CVF Conference on Computer Vision and Pattern Recognition},
  pages={7201--7211},
  year={2022}
}

@inproceedings{song2023ecotta,
  title={Ecotta: Memory-efficient continual test-time adaptation via self-distilled regularization},
  author={Song, Junha and Lee, Jungsoo and Kweon, In So and Choi, Sungha},
  booktitle={Proceedings of the IEEE/CVF Conference on Computer Vision and Pattern Recognition},
  pages={11920--11929},
  year={2023}
}

@inproceedings{niu2022efficient,
  title={Efficient test-time model adaptation without forgetting},
  author={Niu, Shuaicheng and Wu, Jiaxiang and Zhang, Yifan and Chen, Yaofo and Zheng, Shijian and Zhao, Peilin and Tan, Mingkui},
  booktitle={International conference on machine learning},
  pages={16888--16905},
  year={2022},
  organization={PMLR}
}

@article{wang2020tent,
  title={Tent: Fully test-time adaptation by entropy minimization},
  author={Wang, Dequan and Shelhamer, Evan and Liu, Shaoteng and Olshausen, Bruno and Darrell, Trevor},
  journal={arXiv preprint arXiv:2006.10726},
  year={2020}
}

@inproceedings{ma2024improved,
  title={Improved Self-Training for Test-Time Adaptation},
  author={Ma, Jing},
  booktitle={Proceedings of the IEEE/CVF Conference on Computer Vision and Pattern Recognition},
  pages={23701--23710},
  year={2024}
}

@inproceedings{dobler2023robust,
  title={Robust mean teacher for continual and gradual test-time adaptation},
  author={D{\"o}bler, Mario and Marsden, Robert A and Yang, Bin},
  booktitle={Proceedings of the IEEE/CVF Conference on Computer Vision and Pattern Recognition},
  pages={7704--7714},
  year={2023}
}

@inproceedings{wang2023feature,
  title={Feature alignment and uniformity for test time adaptation},
  author={Wang, Shuai and Zhang, Daoan and Yan, Zipei and Zhang, Jianguo and Li, Rui},
  booktitle={Proceedings of the IEEE/CVF Conference on Computer Vision and Pattern Recognition},
  pages={20050--20060},
  year={2023}
}

@inproceedings{wang2023dynamically,
  title={Dynamically instance-guided adaptation: A backward-free approach for test-time domain adaptive semantic segmentation},
  author={Wang, Wei and Zhong, Zhun and Wang, Weijie and Chen, Xi and Ling, Charles and Wang, Boyu and Sebe, Nicu},
  booktitle={Proceedings of the IEEE/CVF Conference on Computer Vision and Pattern Recognition},
  pages={24090--24099},
  year={2023}
}

@inproceedings{kim2022ev,
  title={Ev-tta: Test-time adaptation for event-based object recognition},
  author={Kim, Junho and Hwang, Inwoo and Kim, Young Min},
  booktitle={Proceedings of the IEEE/CVF Conference on Computer Vision and Pattern Recognition},
  pages={17745--17754},
  year={2022}
}

@article{tian2021vdm,
  title={VDM-DA: Virtual domain modeling for source data-free domain adaptation},
  author={Tian, Jiayi and Zhang, Jing and Li, Wen and Xu, Dong},
  journal={IEEE Transactions on Circuits and Systems for Video Technology},
  volume={32},
  number={6},
  pages={3749--3760},
  year={2021},
  publisher={IEEE}
}

@inproceedings{tang2024source,
  title={Source-free domain adaptation with frozen multimodal foundation model},
  author={Tang, Song and Su, Wenxin and Ye, Mao and Zhu, Xiatian},
  booktitle={Proceedings of the IEEE/CVF Conference on Computer Vision and Pattern Recognition},
  pages={23711--23720},
  year={2024}
}

@article{yan2023sam4udass,
  title={Sam4udass: When sam meets unsupervised domain adaptive semantic segmentation in intelligent vehicles},
  author={Yan, Weihao and Qian, Yeqiang and Zhuang, Hanyang and Wang, Chunxiang and Yang, Ming},
  journal={IEEE Transactions on Intelligent Vehicles},
  volume={9},
  number={2},
  pages={3396--3408},
  year={2023},
  publisher={IEEE}
}

@inproceedings{sakaridis2021acdc,
  title={ACDC: The adverse conditions dataset with correspondences for semantic driving scene understanding},
  author={Sakaridis, Christos and Dai, Dengxin and Van Gool, Luc},
  booktitle={Proceedings of the IEEE/ICVF international conference on computer vision},
  pages={10765--10775},
  year={2021}
}

@inproceedings{jaipuria2020deflating,
  title={Deflating dataset bias using synthetic data augmentation},
  author={Jaipuria, Nikita and Zhang, Xianling and Bhasin, Rohan and Arafa, Mayar and Chakravarty, Punarjay and Shrivastava, Shubham and Manglani, Sagar and Murali, Vidya N},
  booktitle={Proceedings of the IEEE/CVF Conference on Computer Vision and Pattern Recognition Workshops},
  pages={772--773},
  year={2020}
}

@article{rohlfs2025generalization,
  title={Generalization in neural networks: A broad survey},
  author={Rohlfs, Chris},
  journal={Neurocomputing},
  volume={611},
  pages={128701},
  year={2025},
  publisher={Elsevier}
}

@article{deng2012mnist,
  title={The mnist database of handwritten digit images for machine learning research},
  author={Deng, Li},
  journal={IEEE Signal Processing Magazine},
  volume={29},
  number={6},
  pages={141--142},
  year={2012},
  publisher={IEEE}
}

@inproceedings{netzer2011reading,
  title={Reading digits in natural images with unsupervised feature learning},
  author={Netzer, Yuval and Wang, Tao and Coates, Adam and Bissacco, Alessandro and Wu, Baolin and Ng, Andrew Y and others},
  booktitle={NIPS workshop on deep learning and unsupervised feature learning},
  volume={2011},
  pages={4},
  year={2011},
  organization={Granada}
}

@inproceedings{radford2021learning,
  title={Learning transferable visual models from natural language supervision},
  author={Radford, Alec and Kim, Jong Wook and Hallacy, Chris and Ramesh, Aditya and Goh, Gabriel and Agarwal, Sandhini and Sastry, Girish and Askell, Amanda and Mishkin, Pamela and Clark, Jack and others},
  booktitle={International conference on machine learning},
  pages={8748--8763},
  year={2021},
  organization={PmLR}
}

@inproceedings{kirillov2023segment,
  title={Segment anything},
  author={Kirillov, Alexander and Mintun, Eric and Ravi, Nikhila and Mao, Hanzi and Rolland, Chloe and Gustafson, Laura and Xiao, Tete and Whitehead, Spencer and Berg, Alexander C and Lo, Wan-Yen and others},
  booktitle={Proceedings of the IEEE/CVF international conference on computer vision},
  pages={4015--4026},
  year={2023}
}

@misc{jocher2023yolov8,
  author = {Glenn Jocher and Ayush Chaurasia and Jing Qiu},
  title = {{YOLO} by Ultralytics},
  year = {2023},
  note = {Available: \url{https://github.com/ultralytics/ultralytics}, accessed 2023}
}

@inproceedings{zhao2023towards,
  title={Towards better stability and adaptability: Improve online self-training for model adaptation in semantic segmentation},
  author={Zhao, Dong and Wang, Shuang and Zang, Qi and Quan, Dou and Ye, Xiutiao and Jiao, Licheng},
  booktitle={Proceedings of the IEEE/CVF conference on computer vision and pattern recognition},
  pages={11733--11743},
  year={2023}
}

@article{zhao2023fast,
  title={Fast segment anything},
  author={Zhao, Xu and Ding, Wenchao and An, Yongqi and Du, Yinglong and Yu, Tao and Li, Min and Tang, Ming and Wang, Jinqiao},
  journal={arXiv preprint arXiv:2306.12156},
  year={2023}
}

@inproceedings{liu2021source,
  title={Source-free domain adaptation for semantic segmentation},
  author={Liu, Yuang and Zhang, Wei and Wang, Jun},
  booktitle={Proceedings of the IEEE/CVF conference on computer vision and pattern recognition},
  pages={1215--1224},
  year={2021}
}

@inproceedings{fleuret2021uncertainty,
  title={Uncertainty reduction for model adaptation in semantic segmentation},
  author={Fleuret, Francois and others},
  booktitle={Proceedings of the IEEE/CVF conference on computer vision and pattern recognition},
  pages={9613--9623},
  year={2021}
}

@inproceedings{richter2016playing,
  title={Playing for data: Ground truth from computer games},
  author={Richter, Stephan R and Vineet, Vibhav and Roth, Stefan and Koltun, Vladlen},
  booktitle={Computer Vision--ECCV 2016: 14th European Conference, Amsterdam, The Netherlands, October 11-14, 2016, Proceedings, Part II 14},
  pages={102--118},
  year={2016},
  organization={Springer}
}

@inproceedings{cordts2016cityscapes,
  title={The cityscapes dataset for semantic urban scene understanding},
  author={Cordts, Marius and Omran, Mohamed and Ramos, Sebastian and Rehfeld, Timo and Enzweiler, Markus and Benenson, Rodrigo and Franke, Uwe and Roth, Stefan and Schiele, Bernt},
  booktitle={Proceedings of the IEEE conference on computer vision and pattern recognition},
  pages={3213--3223},
  year={2016}
}

@inproceedings{chen2019domain,
  title={Domain adaptation for semantic segmentation with maximum squares loss},
  author={Chen, Minghao and Xue, Hongyang and Cai, Deng},
  booktitle={Proceedings of the IEEE/CVF international conference on computer vision},
  pages={2090--2099},
  year={2019}
}

@inproceedings{liu2023multi,
  title={Multi-interactive feature learning and a full-time multi-modality benchmark for image fusion and segmentation},
  author={Liu, Jinyuan and Liu, Zhu and Wu, Guanyao and Ma, Long and Liu, Risheng and Zhong, Wei and Luo, Zhongxuan and Fan, Xin},
  booktitle={Proceedings of the IEEE/CVF international conference on computer vision},
  pages={8115--8124},
  year={2023}
}

@INPROCEEDINGS{10533619,
  author={Varghese, Rejin and M., Sambath},
  booktitle={2024 International Conference on Advances in Data Engineering and Intelligent Computing Systems (ADICS)}, 
  title={YOLOv8: A Novel Object Detection Algorithm with Enhanced Performance and Robustness}, 
  year={2024},
  volume={},
  number={},
  pages={1-6},
  doi={10.1109/ADICS58448.2024.10533619}}

@inproceedings{ye2021source,
  title={Source data-free unsupervised domain adaptation for semantic segmentation},
  author={Ye, Mucong and Zhang, Jing and Ouyang, Jinpeng and Yuan, Ding},
  booktitle={Proceedings of the 29th ACM international conference on multimedia},
  pages={2233--2242},
  year={2021}
}

@article{huang2021model,
  title={Model adaptation: Historical contrastive learning for unsupervised domain adaptation without source data},
  author={Huang, Jiaxing and Guan, Dayan and Xiao, Aoran and Lu, Shijian},
  journal={Advances in neural information processing systems},
  volume={34},
  pages={3635--3649},
  year={2021}
}

@inproceedings{ghiasi2021simple,
  title={Simple copy-paste is a strong data augmentation method for instance segmentation},
  author={Ghiasi, Golnaz and Cui, Yin and Srinivas, Aravind and Qian, Rui and Lin, Tsung-Yi and Cubuk, Ekin D and Le, Quoc V and Zoph, Barret},
  booktitle={Proceedings of the IEEE/CVF conference on computer vision and pattern recognition},
  pages={2918--2928},
  year={2021}
}

@misc{INO2012,
  author       = {{INO}},
  title        = {Video analytics dataset},
  year         = {2012},
  howpublished = {\url{https://www.ino.ca/en/technologies/video-analyticsdataset/}},
  note         = {Accessed: 2025-08-01}
}

@article{li2019rgb,
  title={RGB-T object tracking: Benchmark and baseline},
  author={Li, Chenglong and Liang, Xinyan and Lu, Yijuan and Zhao, Nan and Tang, Jin},
  journal={Pattern Recognition},
  volume={96},
  pages={106977},
  year={2019},
  publisher={Elsevier}
}

@inproceedings{hwang2015multispectral,
  title={Multispectral pedestrian detection: Benchmark dataset and baseline},
  author={Hwang, Soonmin and Park, Jaesik and Kim, Namil and Choi, Yukyung and So Kweon, In},
  booktitle={Proceedings of the IEEE conference on computer vision and pattern recognition},
  pages={1037--1045},
  year={2015}
}

@InProceedings{ji2023mvss,
      title     = {Multispectral Video Semantic Segmentation: A Benchmark Dataset and Baseline},
      author    = {Ji, Wei and Li, Jingjing and Bian, Cheng and Zhou, Zongwei and Zhao, Jiaying and Yuille, Alan L. and Cheng, Li},
      booktitle = {Proceedings of the IEEE/CVF Conference on Computer Vision and Pattern Recognition (CVPR)},
      month     = {June},
      year      = {2023},
      pages     = {1094-1104}
}

@article{chen2017deeplab,
  title={Deeplab: Semantic image segmentation with deep convolutional nets, atrous convolution, and fully connected crfs},
  author={Chen, Liang-Chieh and Papandreou, George and Kokkinos, Iasonas and Murphy, Kevin and Yuille, Alan L},
  journal={IEEE transactions on pattern analysis and machine intelligence},
  volume={40},
  number={4},
  pages={834--848},
  year={2017},
  publisher={IEEE}
}

@article{chen2017rethinking,
  title={Rethinking atrous convolution for semantic image segmentation},
  author={Chen, Liang-Chieh and Papandreou, George and Schroff, Florian and Adam, Hartwig},
  journal={arXiv preprint arXiv:1706.05587},
  year={2017}
}

@inproceedings{he2016deep,
  title={Deep residual learning for image recognition},
  author={He, Kaiming and Zhang, Xiangyu and Ren, Shaoqing and Sun, Jian},
  booktitle={Proceedings of the IEEE conference on computer vision and pattern recognition},
  pages={770--778},
  year={2016}
}
}


\end{document}